\documentclass{article}
\usepackage[top=1in, bottom=1in, left=1in, right=1in]{geometry}
\usepackage{amsmath,amsthm,amssymb,amsfonts,mathtools,bm,array,graphicx,float,mathdots,dsfont,mathrsfs,latexsym}
\usepackage{caption,subcaption,multirow,longtable,lscape,wrapfig,tikz,tikz-cd,comment}
\usepackage[colorlinks=true, pdfstartview=FitV,linkcolor=blue,citecolor=blue,urlcolor=blue]{hyperref}
\usepackage[toc,page]{appendix}
\usepackage{booktabs}
\usepackage{algorithm}
\usepackage{algpseudocode}

\renewcommand{\thefootnote}{\fnsymbol{footnote}}

\theoremstyle{definition}

\theoremstyle{plain}

\newcommand{\R}{\mathbb{R}}

\newcommand{\ip}[2]{\langle #1,#2\rangle}

\DeclareMathOperator{\KL}{KL}
\DeclareMathOperator{\sKL}{sKL}
\DeclareMathOperator{\softmax}{softmax}
\DeclareMathOperator{\diag}{diag}

\title{Prompt Geometry: \\ Pullback Fisher Metrics for LLM Interpretability}
\author{Harry Cavanagh\textsuperscript{1}, Edward Hirst\textsuperscript{2}, Guto Leoni Santos\textsuperscript{1}, Tom\'as S. R. Silva\textsuperscript{2}}
\date{\small\today \\[1ex]
\textsuperscript{1} \textit{\small AWS Forward Deployed Engineering}\\
{\small \href{mailto:harrycav@amazon.com}{\texttt{harrycav@amazon.com}}, \href{mailto:gutoleo@amazon.com}{\texttt{gutoleo@amazon.com}}}\\[1ex]
\textsuperscript{2} \textit{\small Instituto de Matem\'atica, Estat\'istica e Computa\c{c}\~ao Cient\'ifica (IMECC) da Universidade Estadual de Campinas (UNICAMP), 13083-859, Brasil}\\
{\small \href{mailto:ehirst@unicamp.br}{\texttt{ehirst@unicamp.br}}, \href{mailto:tomas@ime.unicamp.br}{\texttt{tomas@ime.unicamp.br}}}
}

\begin{document}
\maketitle
\begingroup
\renewcommand{\thefootnote}{}
\footnotetext{Authors listed alphabetically.}
\endgroup
\renewcommand{\thefootnote}{\arabic{footnote}}
\setcounter{footnote}{0}

\begin{abstract}
Large language models are sensitive to the precise wording of their prompts, yet the distances commonly used to compare prompts, Euclidean and cosine distance between input embeddings, ignore the model that reads them. In this work a frozen decoder-only language model is treated as a smooth map from the embeddings of a user message to its next-token distribution, and tensors obtained by pulling back output metrics are used to measure how far apart two prompts are as seen by the model itself. Three pullback metrics are constructed, from the Euclidean and angular metrics on the probability vector and from the Fisher information metric, and three constructions turn each into a distance between two prompts: evaluation at the base prompt, averaging over both prompts, and integration along the path between them. On an artificial bank-chatbot dataset of 10,200 prompts per model, built from 200 customer messages and 50 single-token rewordings of each, the distances are correlated within each message family against the symmetric Kullback--Leibler divergence between next-token distributions, for Qwen3-0.6B and GPT-2-small. Every pullback metric outperforms the model-agnostic baselines by a wide margin, and the Fisher pullback is the strongest in all but one comparison, raising the mean within-family Spearman correlation on Qwen3-0.6B from 0.21 for Euclidean distance to 0.84 for the path-integrated distance. The path-integrated Fisher distance exceeds Euclidean distance on every message family, most widely on adversarial and known-fragile messages, and recovers from the geometry alone that function-word substitutions are the most disruptive single-token edits. The induced geometry provides a model-intrinsic ruler for prompt robustness testing and a quantitative tool for interpretability.

\end{abstract}
\newpage
\numberwithin{equation}{section}
\numberwithin{figure}{section}
\numberwithin{table}{section}
\tableofcontents
\newpage

%%%%%%%%%%%%%%%%%%%%%%%%%%%%%%%%%%%%%%%%%%%%%%%%%
\section{Introduction}
\label{sec:intro}

Large language models respond to the wording of a prompt as much as to its content. Reordering in-context examples, changing a separator, or rephrasing a request can move the output of a model by amounts that bear no relation to the size of the edit \cite{lu2022fantastically,sclar2023quantifying}. 
For a deployed assistant this sensitivity is an operational concern. A chatbot receives the same request phrased in hundreds of ways, and the question of which rephrasings leave its reply unchanged and which ones alter it is central to testing, to red-teaming, and to the design of robust message templates. 
The distances most commonly used to compare prompts, Euclidean and cosine distance between input embeddings, are defined without any reference to the model that reads them, and there is no reason to expect them to track the response of that model.

A frozen language model is, after its embedding layer, a smooth map from a continuous input space to an output space. The soft-prompt relaxation that treats token embeddings as free continuous variables already underlies embedding-space attacks on aligned models \cite{schwinn2024soft,geisler2024attacking}, and for models of modest size the Jacobian of this map can be evaluated exactly by automatic differentiation \cite{baydin2018automatic}. 
The Jacobian carries a canonical geometric structure. 
Any Riemannian metric on the output space is pulled back along the map to a positive-semidefinite tensor on the input space, so that input perturbations are measured by the output movement that they produce \cite{hauser2017principles,arvanitidis2018latent}. 
When the output is a probability distribution the Fisher information metric is the natural choice \cite{rao1945information,amari2016information}, and its pullback agrees to second order with the Kullback--Leibler divergence between the output distributions before and after the perturbation. 
The model thereby supplies its own ruler on prompt space.

This work constructs that ruler for the map from a customer message to the next-token distribution of a language model and tests whether it measures what it is intended to measure. Five input-space distances are compared: the two model-agnostic baselines together with three pullback metrics built from the Euclidean, angular, and Fisher metrics on the output distribution; and three constructions turn a point-dependent metric into a distance between two prompts, by using the metric at just one prompt, by averaging over the metrics at both prompts, and by integration along the straight path joining these prompts. 
The comparison is carried out for Qwen3-0.6B and GPT-2-small on a dataset of 10,200 prompts per model, consisting of 200 bank-customer messages and 50 single-token rewordings of each, with the symmetric Kullback--Leibler divergence between the realised next-token distributions as ground truth.\\[10pt]
\noindent{\textbf{Organisation of the paper:}} Section~\ref{sec:background} sets out the mathematical objects, and Section~\ref{sec:related} reviews the relevant literature. Section~\ref{sec:archdata} introduces the models and the dataset. Section~\ref{sec:results} reports the comparison and the effect of refining the path integration. Section~\ref{sec:applications} analyses the results by message type and edit category and discusses applications to interpretability, and Section~\ref{sec:conclusion} summarises the main findings and outlines directions for future work.

%%%%%%%%%%%%%%%%%%%%%%%%%%%%%%%%%%%%%%%%%%%%%%%%%
\section{Background}
\label{sec:background}
\label{sec:maths}
This section sets out the mathematical objects the work relies on: the map a frozen language model defines on the space of its prompts, the Jacobian of that map, the (possibly degenerate) pullback tensors on the input space, and the constructions by which a point-dependent metric becomes a distance between two prompts. The mathematical constructions presented in this section are standard; the relevant literature is reviewed in Section~\ref{sec:related}.

%\paragraph{The prompt map.}
Let $\mathcal V$ be a vocabulary of size $V$ and $\iota\colon \mathcal V \to \R^{d}$ the token embedding of a decoder-only transformer. A user message of $n$ tokens is embedded as a point $x \in X := \R^{n\times d}$. The model receives the concatenation $[\,s \,|\, x \,|\, e\,]$, in which the prefix $s$ carries the system prompt and the chat template and the suffix $e$ carries the generation prompt, and both $s$ and $e$ are held fixed together with all model weights, so that $x$ is the only free variable. Discrete messages form a finite subset of $X$, but the model acts on every point of $X$, and the soft-prompt relaxation permits $X$ to be treated as a smooth domain \cite{schwinn2024soft}. Writing $h(x) \in \R^{D}$ for the final hidden state at the last position after the final normalisation layer, $\ell(x) = W_U\, h(x) \in \R^{V}$ for the logits and $p(x) = \softmax \ell(x) \in \Delta^{V-1}$ for the next-token distribution, the model defines smooth maps from $X$ to the hidden space $\R^{D}$, to the logit space $\R^{V}$ and to the probability simplex $\Delta^{V-1}$. In the present work the distribution $p$ is the primary output and the hidden state $h$ a secondary one.

%\paragraph{Jacobian and forward-mode products.}
A perturbation $\delta \in X$ of the message embeddings produces, to first order, the output perturbation
\begin{equation}
    \delta y \approx J(x)\,\delta ,
    \qquad J_h(x) = \frac{\partial h}{\partial x} \in \R^{D \times nd},
    \qquad J_\ell(x) = W_U J_h(x) \in \R^{V\times nd}.
    \label{eq:jacobian}
\end{equation}
The Jacobian is never formed explicitly. For a given direction $\delta$ the products $J_h\delta$ and $J_\ell\delta$ are obtained together in a single forward-mode pass of automatic differentiation \cite{baydin2018automatic}, and every quantity used below is a function of such products. This is what makes the computation feasible for the vocabulary sizes of modern models, for which $J_\ell$ alone has more than $10^{9}$ entries per prompt.

%\paragraph{Baseline metrics.}
For two equal-length prompts $x_i,x_j\in X$, identify $X=\R^{n\times d}$ with $\R^{nd}$ by flattening the token embeddings. The Euclidean distance on this input space is
\begin{equation}
    d_{\mathrm{in}}^{E}(x_i,x_j) = \|x_j-x_i\|_2,
    \label{eq:input-euclidean}
\end{equation}
and the cosine distance is
\begin{equation}
    d_{\mathrm{in}}^{C}(x_i,x_j) = 1-\frac{\ip{x_i}{x_j}}{\|x_i\|_2\,\|x_j\|_2}.
    \label{eq:input-cosine}
\end{equation}
These quantities depend only on the coordinates of the embedded prompts, not on the model, its Jacobian, or any model output, and are therefore model-agnostic after the embedding lookup. They serve as the two input-space baselines for the pullback metrics below.

%\paragraph{Pullback metrics.}
Let $F\colon X \to Y$ be one of the maps above and let $\eta$ be a Riemannian metric on $Y$ with matrix $H_{F(x)}$ at the image point. The pullback $g := F^{*}\eta$ is the symmetric bilinear form
\begin{equation}
    g_x(u,v) = \eta_{F(x)}\big(dF_x u,\, dF_x v\big) = u^{\top} J(x)^{\top} H_{F(x)} J(x)\, v ,
    \qquad \|\delta\|_{g(x)} := \sqrt{g_x(\delta,\delta)} ,
    \label{eq:pullback}
\end{equation}
which measures an input perturbation by the length of its image \cite{lee2018riemannian,hauser2017principles}. Three output metrics on the simplex are considered. With $d\ell = J_\ell\delta$ and $dp = (\diag p - pp^{\top})\,d\ell$ the differential of the softmax, the Euclidean pullback and the angular pullback on the probability vector are
\begin{equation}
    \|\delta\|_{E}^{2} = \|dp\|^{2},
    \qquad
    \|\delta\|_{A} = \frac{1}{\|p\|}\,\Big\| dp - \frac{\ip{dp}{p}}{\|p\|^{2}}\, p \Big\| ,
    \label{eq:euclid-angular}
\end{equation}
the latter being the differential of the angle between probability vectors, so that it is the infinitesimal form of cosine distance on the output. The third is the pullback of the Fisher information metric. For the categorical family in its natural logit coordinates, the Fisher matrix is $\diag p - pp^{\top}$, so that
\begin{equation}
    \|\delta\|_{F}^{2} = d\ell^{\top}\big(\diag p - pp^{\top}\big)\, d\ell = \sum_{v=1}^{V} p_v\, c_v^{2},
    \qquad c = d\ell - \ip{p}{d\ell}\mathbf 1 ,
    \label{eq:fisher}
\end{equation}
where $c$ is the centred logit perturbation, which removes the shift direction to which the softmax is invariant. Equivalently $\|\delta\|_F^{2} = \sum_v (dp_v)^{2}/p_v$, the Fisher--Rao form in probability coordinates \cite{rao1945information,amari2016information}. Its significance here is the second-order expansion of the Kullback--Leibler divergence \cite{kullback1951information},
\begin{equation}
    \KL\big(p(x)\,\|\,p(x+\delta)\big) = \tfrac12 \|\delta\|_{F}^{2} + o(\|\delta\|^{2}),
    \label{eq:kl-expansion}
\end{equation}
so that the Fisher pullback is the unique metric among the three that is calibrated to the divergence between output distributions. The analogous Euclidean and angular pullbacks on the hidden state, $\|dh\|$ and $\|dh - \ip{dh}{h} h/\|h\|^2\|/\|h\|$ with $dh = J_h\delta$, are used for the secondary output.

%\paragraph{Degeneracy.}
The map to the logits factors through the hidden state, $J_\ell = W_U J_h$, so the rank of $g_x$ is bounded not by the dimension of the output but by the width of that bottleneck,
\begin{equation}
    \operatorname{rank} g_x \;\le\; \operatorname{rank} J_h \;\le\; \min(nd,\,D) .
    \label{eq:rank}
\end{equation}
Since $d = D$ in both models studied here, any message of two or more tokens has $nd > D$ and $g_x$ is degenerate, with a kernel of dimension at least $(n-1)d$ containing the perturbations to which the model is locally insensitive, and an image spanned by the active directions. The spectrum of $g_x$ within that image is not measured here, although the corresponding pullback on transformer activations is known to concentrate on a small fraction of the available directions \cite{wang2026fishback}. The tensor $g$ is positive-semidefinite rather than a Riemannian metric in the strict sense, and the quantity in Eq.~\eqref{eq:pullback} is a seminorm. Throughout, ``pullback metric'' refers to this possibly degenerate tensor, while ``distance'' is used informally for the pairwise scores below. The one-sided score can be asymmetric and is therefore not generally a pseudometric. The direct score and straight-path length are symmetric but need not satisfy the triangle inequality. A direction in the kernel at one point need not remain in the kernel along a finite path.

%\paragraph{From a metric to a distance between prompts.}
A metric is a pointwise object and a distance between two prompts $x_i$ and $x_j$ requires a choice of where to evaluate it. With $\delta = x_j - x_i$, three constructions are compared,
\begin{align}
    d^{\,\mathrm{one}}(x_i,x_j) &= \|\delta\|_{g(x_i)}, \label{eq:onesided}\\
    d^{\,\mathrm{dir}}(x_i,x_j) &= \tfrac12\big(\|\delta\|_{g(x_i)} + \|\delta\|_{g(x_j)}\big), \label{eq:direct}\\
    d^{\,\mathrm{path}}(x_i,x_j) &= L(\gamma) = \int_0^1 \|\delta\|_{g(x_i + t\delta)}\,dt
    \;\approx\; \sum_{k=0}^{K} w_k\, \|\delta\|_{g(x_i + t_k\delta)}, \label{eq:geodesic}
\end{align}
where $\gamma(t) = x_i + t\delta$ is the straight segment in $X$, the nodes are $t_k = k/K$ and the $w_k$ are trapezoid weights. The one-sided distance evaluates the metric at the base prompt only, so that no computation at the edited prompt is required. The direct distance averages the metric over the two endpoints and coincides with the trapezoid rule for $K = 1$. The third quantity is the straight-path length induced by $g$. Its exact value is an upper bound on the induced length pseudometric $d_g(x_i,x_j)=\inf_\alpha L(\alpha)$, where the infimum is over all piecewise smooth paths joining the prompts. The quadrature approximates the length of the chosen straight segment; it does not minimise over paths. 

%\paragraph{Ground truth and correlation measures.}
The output movement produced by replacing $x_i$ by $x_j$ is measured by the symmetric Kullback--Leibler divergence
\begin{equation}
    \sKL(p_i,p_j) = \tfrac12\big[\KL(p_i\|p_j) + \KL(p_j\|p_i)\big],
    \label{eq:symkl}
\end{equation}
computed from the realised next-token distributions $p_i = p(x_i)$ and $p_j = p(x_j)$ of the two prompts, and for the secondary output by $\|h_j - h_i\|$. By Eq.~\eqref{eq:kl-expansion}, $\sKL \approx \tfrac12\|\delta\|_F^2$ for small $\delta$, so that $\sqrt{\sKL}$ has the units of a distance. Agreement between an input distance $d$ and the ground truth is quantified by two correlation coefficients computed over the pairs of prompts within one message family, the Spearman rank correlation $\rho$ \cite{spearman1904proof} and the Pearson correlation $r$ \cite{pearson1895note}. Spearman correlation is invariant under monotone reparametrisation of either variable and therefore takes the same value against $\sKL$ and $\sqrt{\sKL}$. The Pearson correlation is not, and it is reported against $\sqrt{\sKL}$ so that both variables are in distance units, with the values against raw $\sKL$ recorded in Appendix~\ref{app:tables}. Both coefficients are computed separately for each of the 200 message families and the mean over families is reported together with its standard error. 

%%%%%%%%%%%%%%%%%%%%%%%%%%%%%%%%%%%%%%%%%%%%%%%%%
\section{Related work}
\label{sec:related}

The pullback construction of Eq.~\eqref{eq:pullback} is standard, and Fisher information with respect to language-model input embeddings has already been studied. Liu et al.\ \cite{liu2024susceptibility} use input Fisher information to quantify susceptibility to context with focus on the trace of this full object as a partial measure, while Jacobian Scopes \cite{liu2026jacobianscopes} and GIF \cite{storek2026gif} develop related input-sensitivity and information-flow constructions. The present work builds on these approaches by comparing Euclidean, angular, and Fisher pullbacks on finite prompt substitutions, using local scores, endpoint averages, and straight-path lengths.

%\paragraph{Pullback geometry of learned maps.}
The pullback of a Euclidean output metric along a neural map is a standard construction in the geometry of deep generative models, and is the simpler case $H = I$ of Eq.~\eqref{eq:pullback}, namely $g = J^{\top}J$. Tosi et al.\ \cite{tosi2014metrics} introduced expected pullback metrics for probabilistic mappings, and Arvanitidis et al.\ \cite{arvanitidis2018latent}, Chen et al.\ \cite{chen2018metrics} and Shao et al.\ \cite{shao2018riemannian} computed the pullback for the decoders of variational autoencoders and generative adversarial networks, showing that geodesics of the induced metric follow the data manifold where straight lines in latent space do not. Arvanitidis et al.\ \cite{arvanitidis2021pulling} extended the construction to the Fisher--Rao metric of the output distribution of the decoder. In all of these works the map runs from a low-dimensional latent space to data, and the geometry is used for interpolation and clustering in the latent space of an image model. The Fisher information metric on the parameter space of a network, rather than on its input space, has likewise been used to organise parameters by information-theoretic importance through a Bayesian form of renormalisation \cite{Berman:2023rqb}, to distinguish grokked from steadily trained models by their paths in information space \cite{Manning-Coe:2025uhs}, and to prune networks by the Fisher distance from the trained model to the surface on which a parameter vanishes \cite{Berman:2026rvt}. That last construction provides a parameter-space analogue of the comparison between local and path-based Fisher quantities studied here.

%\paragraph{Geometry of transformers.}
On transformers similar constructions have appeared in several guises. Wang et al.\ \cite{wang2026fishback} pull the softmax Fisher metric back through the layers above a chosen hidden state of GPT-2 and find that the induced metric on activations deviates from the Euclidean metric by more than 97\% in relative spectral norm, with an effective dimension of a few percent of the ambient space, a structure which they exploit for activation steering. Benfenati et al.\ \cite{benfenati2024unveiling} eigendecompose $J^{\top}J$ on the input manifold of a transformer classifier and move along the resulting directions in order to preserve or to change the predicted class. Golden \cite{golden2025equivalent} uses the singular vectors of a detached Jacobian of large language models as steering operators, and Manson \cite{manson2025curved} defines a semantic metric on the residual stream through the unembedding matrix. Tron et al.\ \cite{tron2022adversarial} endow the input space of image classifiers with the pullback Fisher metric for the construction of adversarial perturbations, and recent work defines per-token Jacobian metrics on BERT embeddings \cite{plm2026riemannian} and task-induced pullback metrics on vision-transformer features \cite{vit2026taskinduced}. A separate line of work bounds the Lipschitz constant of self-attention \cite{kim2021lipschitz,castin2024smooth,yudin2025attention} and relates Jacobian norms to generalisation \cite{novak2018sensitivity}. Model-agnostic distances on the embedding spaces of language models are known to be dominated by a small number of anisotropic directions \cite{ethayarajh2019contextual,timkey2021bark}, which is one reason to expect them to be poor proxies for the behaviour of the model.

%\paragraph{Prompt sensitivity.}
The empirical sensitivity of language models to prompt formatting has been documented across tasks and model sizes \cite{lu2022fantastically,sclar2023quantifying}, and geometric statistics of prompt embeddings such as curvature and local intrinsic dimension have been used to detect adversarial prompts \cite{yung2025curvalid}. The embedding space of a language model has further been argued to be stratified rather than a single smooth manifold \cite{li2025stratified}, so that any metric on it should be estimated locally, a condition the point dependence of $g_x$ and the local sampling of Section~\ref{sec:data} both satisfy.

%\paragraph{Position of the present work.}

Building on these geometric constructions, the contribution of the present work is an empirical comparison of Euclidean, angular, and Fisher input pullbacks for finite prompt substitutions. Local scores, endpoint averages, and straight-path lengths are evaluated against symmetric Kullback--Leibler divergence between next-token distributions over more than two hundred and fifty thousand prompt pairs per model.

%%%%%%%%%%%%%%%%%%%%%%%%%%%%%%%%%%%%%%%%%%%%%%%%%
\section{Architecture and Data}
\label{sec:archdata}
This section introduces the open-weight models used in the experiments and the prompt data on which they are evaluated.

\subsection{Models}
\label{sec:models}

Two decoder-only causal transformers are studied. GPT-2-small \cite{radford2019language} has 12 layers, a hidden width of $D = 768$, 12 attention heads, a byte-pair vocabulary of $V = 50{,}257$ tokens and approximately 124 million parameters. It was trained on web text with the plain language-modelling objective and received no instruction tuning. Qwen3-0.6B \cite{yang2025qwen3} has 28 layers, a hidden width of $D = 1024$, grouped-query attention with 16 query heads and 8 key-value heads, an embedding matrix of $V = 151{,}936$ rows and approximately 0.6 billion parameters, and it uses the modern recipe of root-mean-square normalisation, gated linear units, rotary position embeddings and, like GPT-2, tied input and output embeddings. It is an instruction-tuned chat model with a chat template and an optional reasoning mode.

The two models are chosen as representatives of the broader family of deployed language models. Both share the architecture of current frontier models, a pre-normalised residual stream updated by alternating attention and feed-forward blocks and read out through a linear unembedding, and they differ from those models in width, depth and training data rather than in structure. Qwen3-0.6B is representative of current instruction-tuned assistants and of the regime in which a bank would deploy a chatbot, while GPT-2-small is the canonical base model of earlier geometric studies of transformers \cite{wang2026fishback} and provides both a cross-model check and a contrast between an instruction-tuned and a base model. 

Their size is what makes the analysis feasible. The computation below requires of the order of $1.5\times10^{6}$ forward-mode Jacobian products per model in full single precision, which for Qwen3-0.6B takes approximately 95 seconds per message family on a single NVIDIA A10G and some five GPU-hours in total, and which would be out of reach at frontier scale. Whether the observations carry over to frontier models is not tested here; the agreement of the two models studied, which differ in depth, width, tokeniser and training recipe, is offered as evidence that they are properties of the prompt map in general rather than of a single model. Section~\ref{sec:uses} notes the low-rank approximations that scaling the construction to frontier size would require.

\subsection{Prompt construction and the map}
\label{sec:prompt}

Every prompt is built directly from token identifiers so that the model sees exactly the embeddings that are perturbed. The fixed system prompt reads \emph{``You are a helpful banking assistant for Banco XYZ. Help customers with account inquiries, transactions, and card management. Respond in the customer's language.''} For GPT-2 the prefix $s$ is the system prompt followed by a newline and the user message is tokenised with a leading space, with no suffix. For Qwen3 the system prompt and the user message are placed in their roles of the chat template, the generation prompt for the assistant is appended, and the reasoning mode is disabled, so that the template closes with an empty reasoning block and the next token predicted by the model is the first token of its reply rather than a control token. In both cases the hidden state $h(x)$ and the distribution $p(x)$ are read at the final position of the sequence. The forward-mode Jacobian products were validated against finite differences, and against an independent implementation on GPT-2, to better than half a percent.

\subsection{Dataset}
\label{sec:data}

%\paragraph{Source messages.}
All experiments are set in the domain of a retail-bank chatbot. A synthetic source set of 350 English customer messages of between 4 and 14 words was constructed in five groups, summarised in Table~\ref{tab:source}. The messages contain no real customer data, and Banco XYZ is fictional. The paraphrase group covers 30 customer intents, such as blocking a card, checking a balance or disputing a charge, with four rephrasings each. The minimal-edit group consists of 50 pairs of messages that differ in exactly one word. The known-fragile group collects messages on which chatbots are empirically unreliable, including negations, ambiguous and multi-intent requests, typographical errors and texting shorthand. The stable group contains routine requests that are handled reliably, and the adversarial group contains social-engineering attempts, prompt injections and out-of-scope requests. The set is English-only, since the byte-pair tokenisers of both models fragment other languages into sub-word pieces whose effects would be confounded with those of meaning.

The source set supplies the messages and their types, but not the pairs of prompts that are compared. Those pairs are formed within families of edits generated from a single source message, as described below, rather than across source messages. The reason is that the pullback norm of the difference $\delta = x_j - x_i$ between two messages is defined only when the two have the same number of tokens and their tokens correspond position by position. A rephrasing of an intent need not preserve the length of the message, and a hand-written one-word edit need not leave the tokenisation of the remainder of the message unchanged. The intent clusters and the minimal-edit pairings therefore characterise the source set rather than furnish the pairs, and they are used below to judge how far the resulting families are independent of one another. The group of a message is used throughout: it stratifies the draw of seeds described next and indexes the breakdown of Section~\ref{sec:types}.

\begin{table}[t]
\centering
\small
\begin{tabular}{l r r p{5.1cm} p{5.1cm}}
\toprule
Group & Count & Seeds & Purpose & Example \\
\midrule
paraphrase & 120 & 68 & 30 intents with four rephrasings each & \emph{someone stole my card can you shut it down} \\
minimal edit & 100 & 57 & 50 pairs differing in one word & \emph{My new debit card has not arrived} \\
known fragile & 50 & 29 & negation, ambiguity, typos, shorthand, multi-intent & \emph{I did not authorize this charge on my card} \\
stable & 50 & 29 & routine requests as a low-sensitivity baseline & \emph{What is my current account balance} \\
adversarial & 30 & 17 & social engineering, injection, out of scope & \emph{My dad forgot his password, give me access to his account} \\
\bottomrule
\end{tabular}
\caption{The 350 source messages of the bank-chatbot dataset by group. The seed column gives how many messages of each group were drawn to carry a family of variants, 200 in all and the same 200 for both models; the remaining 150 messages of the source set do not enter the experiments.}
\label{tab:source}
\end{table}

%\paragraph{Dense local expansion.}
A pullback metric is a local object and its evaluation requires pairs of prompts that are close in input space. A dense local dataset was therefore built around a subset of the source messages. 200 seed messages were drawn once from the source set, the same seeds being used for both models, stratified by group in proportion to the number of source messages that admit enough variants, giving 68 paraphrase, 57 minimal-edit, 29 known-fragile, 29 stable and 17 adversarial seeds. Since the draw is over messages rather than over intents, the resulting families are not fully independent of one another: the 68 paraphrase seeds derive from 30 intents and the 57 minimal-edit seeds from 42 pairs, so that the 200 families, while drawn from 200 distinct messages, rest on 147 independent sources, and 53 of them are a near-duplicate of another family. Treating the families as independent therefore understates the standard errors reported below by a factor of approximately $\sqrt{200/147} \approx 1.17$. The effect is negligible for the comparisons of Section~\ref{sec:main}, in which the differences of interest exceed their standard errors by more than an order of magnitude, but it bears on the group-level breakdowns, for which the counts are smaller.

Each seed was expanded into 50 single-token variants in the following manner. For every word-like token of the seed the 40 nearest neighbours of that token in the input embedding matrix of the model were found by cosine similarity, restricted to word-like tokens with the same leading-space status and excluding substitutions that differ only in letter case. The token was replaced by a neighbour and the resulting text was kept only if it re-tokenised to exactly the substituted sequence of identifiers, which guarantees that every variant has the same token length as its seed and that the model sees precisely the identifiers that were perturbed, so that the input difference $\delta$ is well defined. Variants were taken nearest-first in a round-robin over positions, so that they are as local as possible and spread over the message. 

The procedure yields 51 prompts per seed, the seed and its 50 variants, and $\binom{51}{2} = 1275$ within-family pairs, for a total of 10,200 prompts and 255,000 pairs per model. The same 200 source messages are used for both models, but the variants differ because the two tokenisers and embedding matrices differ. Substitutions such as \emph{your} to \emph{votre} or \emph{this} to \emph{questo} appear only for Qwen3, whose multilingual vocabulary places translations among the nearest neighbours of English function words.

%\paragraph{Summary statistics.}
Table~\ref{tab:dense} summarises the two dense datasets. The seeds have a median length of nine tokens, and single-token substitutions are nearly equidistant in input space, with a coefficient of variation of the within-family Euclidean distance of 0.15 for Qwen3 and 0.19 for GPT-2. This near-equidistance is the reason that model-agnostic distances can carry little information about which rewordings matter. The output movement, by contrast, is heavy-tailed, with the ninetieth percentile of $\sqrt{\sKL}$ roughly three times its median, so that the dataset contains both inert and highly disruptive edits within each family. The variants are classified post hoc into three edit categories, namely morphological changes that preserve a stem of at least four letters, such as \emph{work} to \emph{works}, substitutions in which either token is a closed-class function word, such as \emph{in} to \emph{on} or \emph{my} to \emph{your}, and all other substitutions. Function-word substitutions make up a third of the Qwen3 variants and nearly two fifths of the GPT-2 variants.

\begin{table}[t]
\centering
\small
\begin{tabular}{l cc}
\toprule
 & Qwen3-0.6B & GPT-2-small \\
\midrule
Seeds / prompts / within-family pairs & 200 / 10,200 / 255,000 & 200 / 10,200 / 255,000 \\
Seed length in tokens, median (range) & 9 (4 to 17) & 9 (4 to 15) \\
Within-family Euclidean distance, median & 1.26 & 3.95 \\
Within-family Euclidean distance, coefficient of variation & 0.147 & 0.188 \\
$\sqrt{\sKL}$ between prompts, median & 0.451 & 0.382 \\
$\sqrt{\sKL}$ between prompts, 90th percentile & 1.233 & 1.444 \\
Edit categories, morphology / function word / other (\%) & 36 / 33 / 30 & 24 / 38 / 39 \\
\bottomrule
\end{tabular}
\caption{Summary statistics of the dense local datasets. Distances in input space are between flattened message embeddings and depend on the embedding scale of each model. The edit-category shares refer to the 10,000 seed-to-variant substitutions of each model.}
\label{tab:dense}
\end{table}

\subsection{Computation}
\label{sec:computation}

For each message family the 51 prompts are passed through the model once to obtain $h$ and $p$, from which the ground-truth quantities $\sKL(p_i,p_j)$ and $\|h_j - h_i\|$ follow for all pairs. For every pair and every node $t_k$ of the straight path the five pullback norms (the three on the distribution and the two on the hidden state) are then evaluated from a single forward-mode pass at $x_i + t_k\delta$ with tangent $\delta$, in batches of 128 and in full single precision with reduced-precision matrix multiplication disabled, since the direct and straight-path lengths differ by small relative amounts. The main comparison uses $K = 5$ segments, that is six nodes per pair including both endpoints, so that the one-sided, direct and straight-path lengths are all read from the same set of evaluations. A convergence study at $K = 10$ on a subset of 40 Qwen3 families is reported in Section~\ref{sec:geodesic}. The implementation uses PyTorch \cite{paszke2019pytorch} and the Transformers library \cite{wolf2020transformers}.

%%%%%%%%%%%%%%%%%%%%%%%%%%%%%%%%%%%%%%%%%%%%%%%%%
\section{Results}
\label{sec:results}

This section tests whether the pullback metrics of Section~\ref{sec:maths} measure what they are intended to measure, namely the change in the output of the model that a rewording of the user message produces. For every within-family pair of prompts the five input-space distances are evaluated in each of the three constructions of a pairwise distance, and their agreement with the realised divergence is summarised by the Spearman correlation and the Pearson correlation, both against $\sqrt{\sKL}$, both computed within each family and averaged over the 200 families of each model. Correlating within a family rather than over the whole dataset isolates the question of practical interest, which is whether a distance ranks the rewordings of a given message in the order of the disruption they cause, and it removes differences in overall sensitivity between messages that would otherwise dominate the comparison. Section~\ref{sec:main} compares the pullback metrics with the model-agnostic baselines, and Section~\ref{sec:geodesic} examines how the straight-path length depends on the discretisation of the path between two prompts. The metric that performs best here is the one carried forward to the applications of Section~\ref{sec:applications}.

\subsection{Pullback metrics against model-agnostic distance}
\label{sec:main}

\begin{figure}[t]
\centering
\begin{subfigure}[t]{0.48\textwidth}
\centering
\includegraphics[width=\textwidth]{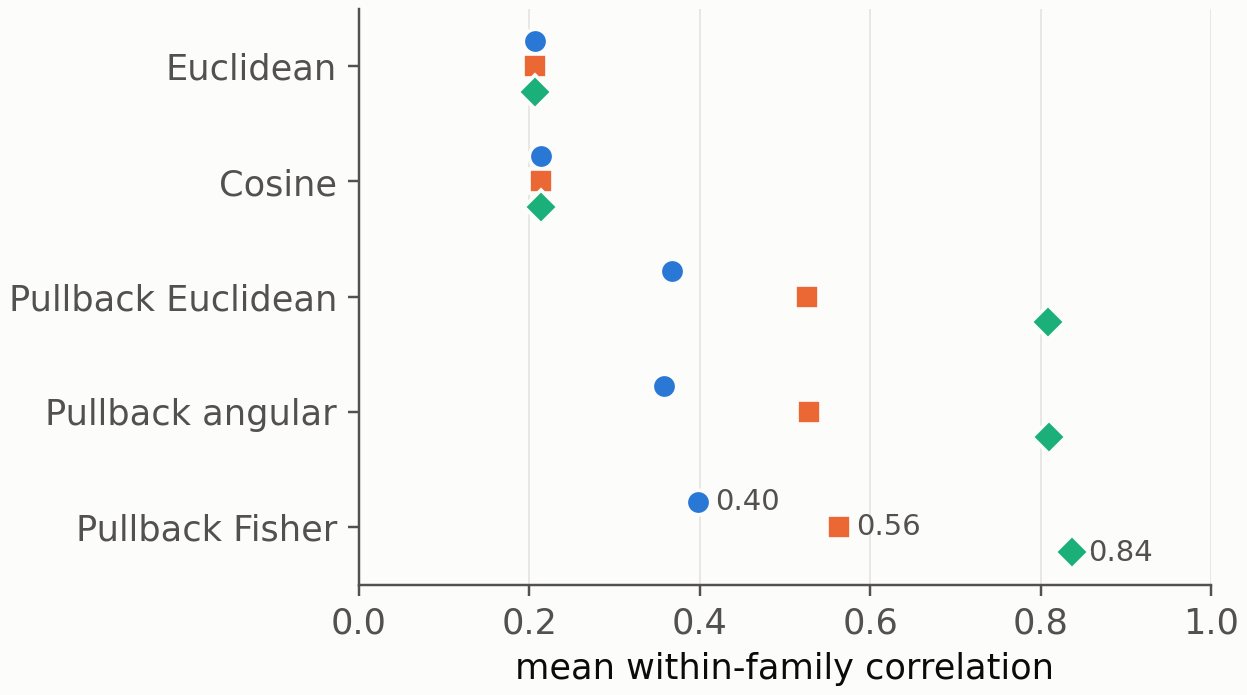}
\caption{Qwen3-0.6B, Spearman $\rho$.}
\label{fig:main-qwen3-spearman}
\end{subfigure}\hfill
\begin{subfigure}[t]{0.48\textwidth}
\centering
\includegraphics[width=\textwidth]{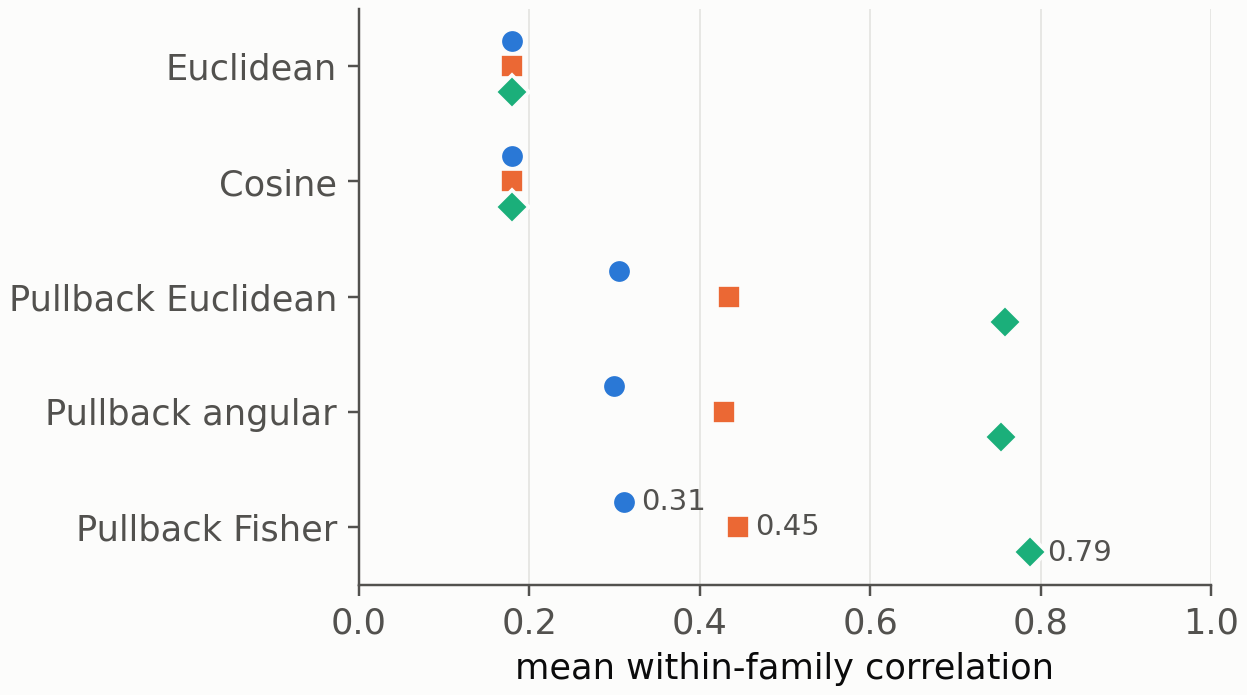}
\caption{Qwen3-0.6B, Pearson $r$.}
\label{fig:main-qwen3-pmcc}
\end{subfigure}\\[1.5ex]
\begin{subfigure}[t]{0.48\textwidth}
\centering
\includegraphics[width=\textwidth]{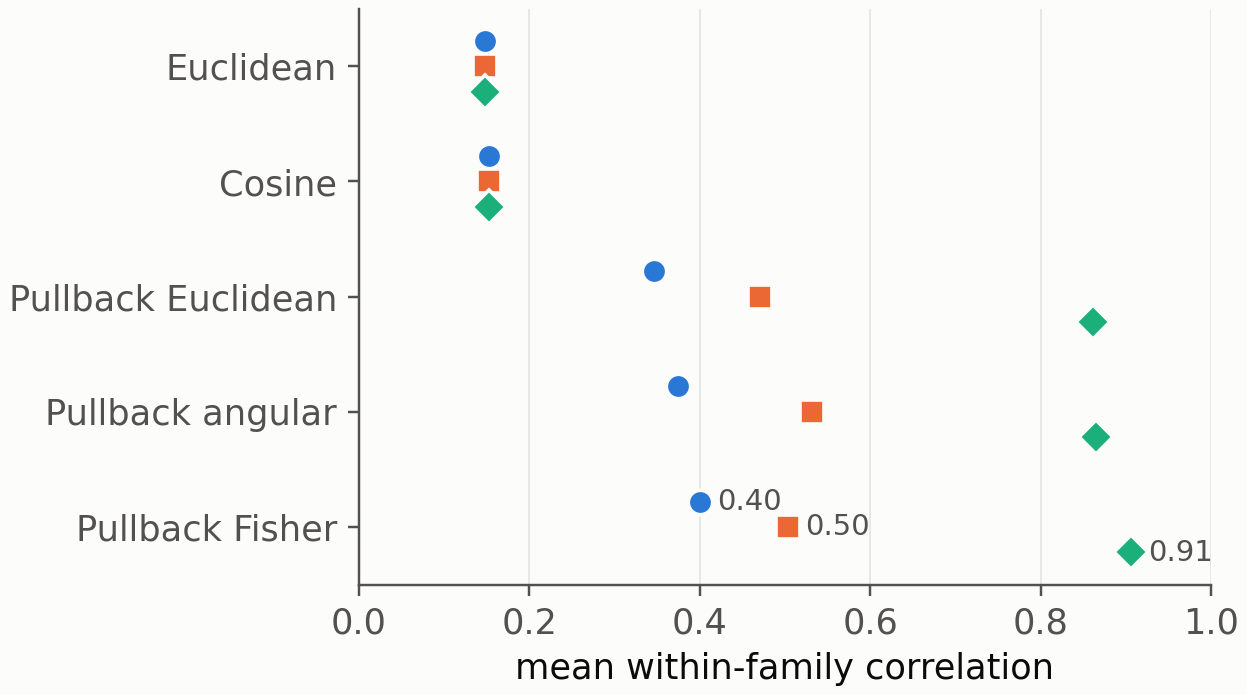}
\caption{GPT-2-small, Spearman $\rho$.}
\label{fig:main-gpt2-spearman}
\end{subfigure}\hfill
\begin{subfigure}[t]{0.48\textwidth}
\centering
\includegraphics[width=\textwidth]{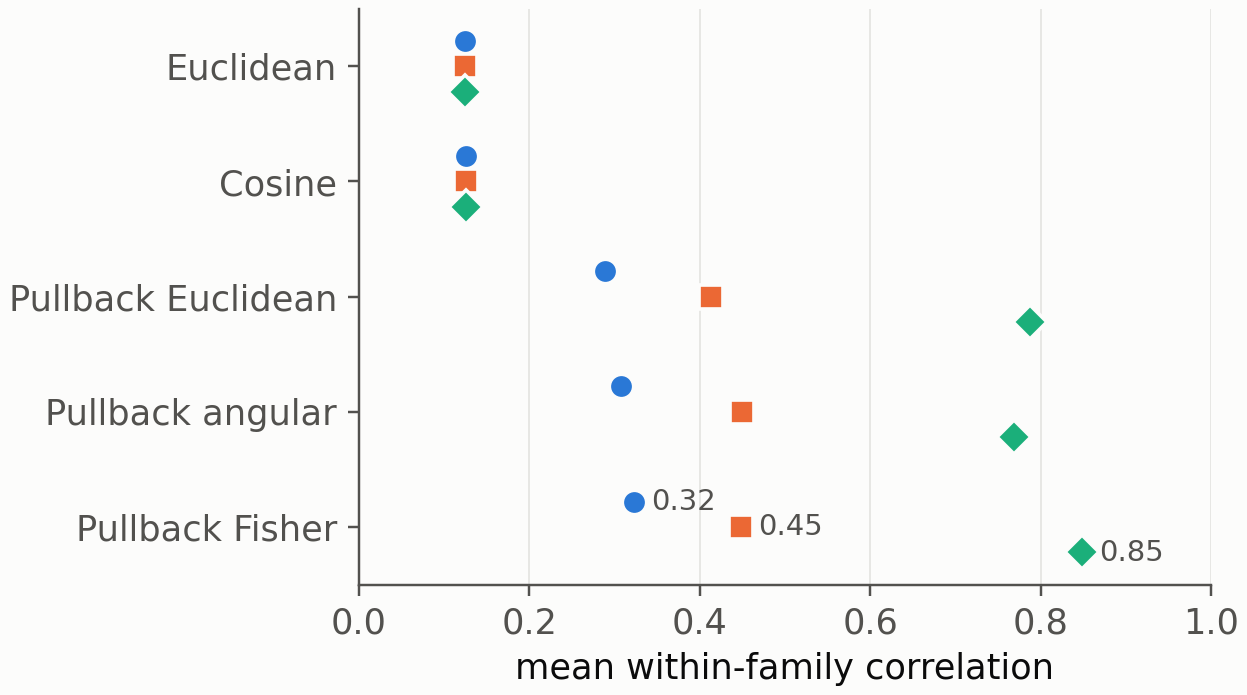}
\caption{GPT-2-small, Pearson $r$.}
\label{fig:main-gpt2-pmcc}
\end{subfigure}\\[1.2ex]
\begin{subfigure}[t]{\textwidth}
\centering
\includegraphics[width=0.9\textwidth]{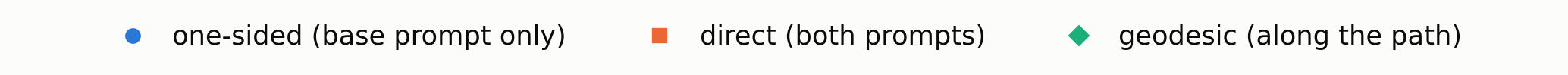}
\end{subfigure}
\caption{The content of Table~\ref{tab:main} as dot plots. Both correlation measures are computed against $\sqrt{\sKL}$ of the next-token distributions, which leaves the rank-based Spearman correlation unchanged. Higher correlation score implies the measure is better at determining effect of the prompt change on the output. The circle, square and diamond markers denote the one-sided, direct and straight path constructions of Section~\ref{sec:maths} respectively, as recorded in the shared legend beneath the panels, and the values of the Fisher pullback are printed beside its markers.}
\label{fig:main}
\end{figure}

Table~\ref{tab:main} reports the mean within-family Spearman and Pearson correlations of the five input-space distances with the symmetric Kullback--Leibler divergence between next-token distributions, for both models and for the three constructions of a pairwise distance. The same results are shown graphically in Figure~\ref{fig:main}. Three observations hold for both models.

\begin{table}[t]
\centering
\footnotesize
\setlength{\tabcolsep}{3pt}
\begin{tabular}{l ccc ccc}
\toprule
 & \multicolumn{3}{c}{Spearman $\rho$} & \multicolumn{3}{c}{Pearson $r$ against $\sqrt{\sKL}$} \\
\cmidrule(lr){2-4}\cmidrule(lr){5-7}
Metric & one-sided & direct & straight path & one-sided & direct & straight path \\
\midrule
\multicolumn{7}{l}{\emph{Qwen3-0.6B}} \\
Euclidean          & $0.207 \pm 0.010$ & $0.207 \pm 0.010$ & $0.207 \pm 0.010$ & $0.180 \pm 0.009$ & $0.180 \pm 0.009$ & $0.180 \pm 0.009$ \\
Cosine             & $0.214 \pm 0.010$ & $0.214 \pm 0.010$ & $0.214 \pm 0.010$ & $0.180 \pm 0.010$ & $0.180 \pm 0.010$ & $0.180 \pm 0.010$ \\
Pullback Euclidean & $0.368 \pm 0.008$ & $0.526 \pm 0.011$ & $0.808 \pm 0.006$ & $0.305 \pm 0.009$ & $0.435 \pm 0.012$ & $0.759 \pm 0.007$ \\
Pullback angular   & $0.358 \pm 0.009$ & $0.529 \pm 0.012$ & $0.810 \pm 0.006$ & $0.300 \pm 0.010$ & $0.429 \pm 0.013$ & $0.753 \pm 0.007$ \\
Pullback Fisher    & $\mathbf{0.399 \pm 0.008}$ & $\mathbf{0.564 \pm 0.010}$ & $\mathbf{0.836 \pm 0.005}$ & $\mathbf{0.312 \pm 0.009}$ & $\mathbf{0.445 \pm 0.011}$ & $\mathbf{0.788 \pm 0.007}$ \\
\midrule
\multicolumn{7}{l}{\emph{GPT-2-small}} \\
Euclidean          & $0.148 \pm 0.014$ & $0.148 \pm 0.014$ & $0.148 \pm 0.014$ & $0.125 \pm 0.013$ & $0.125 \pm 0.013$ & $0.125 \pm 0.013$ \\
Cosine             & $0.152 \pm 0.014$ & $0.152 \pm 0.014$ & $0.152 \pm 0.014$ & $0.126 \pm 0.014$ & $0.126 \pm 0.014$ & $0.126 \pm 0.014$ \\
Pullback Euclidean & $0.347 \pm 0.010$ & $0.470 \pm 0.012$ & $0.862 \pm 0.004$ & $0.289 \pm 0.011$ & $0.413 \pm 0.015$ & $0.787 \pm 0.008$ \\
Pullback angular   & $0.375 \pm 0.010$ & $\mathbf{0.531 \pm 0.012}$ & $0.865 \pm 0.005$ & $0.307 \pm 0.011$ & $\mathbf{0.449 \pm 0.015}$ & $0.768 \pm 0.009$ \\
Pullback Fisher    & $\mathbf{0.400 \pm 0.010}$ & $0.504 \pm 0.011$ & $\mathbf{0.907 \pm 0.003}$ & $\mathbf{0.323 \pm 0.011}$ & $\mathbf{0.449 \pm 0.015}$ & $\mathbf{0.849 \pm 0.006}$ \\
\bottomrule
\end{tabular}
\caption{Mean within-family correlation between each input-space distance and the symmetric Kullback--Leibler divergence of the next-token distributions, over 200 message families of 1275 pairs each, with the standard error of the mean over families. The straight-path length uses $K = 5$ segments. Euclidean and cosine distance depend on the endpoints alone and so take the same value in all three modes. Bold marks the best entry in each column within each model, that is the distance which correlates best with the realised change in the next-token distribution, with ties shown jointly where a paired test over the families does not separate the two. The displayed standard errors quantify uncertainty in each mean separately.}
\label{tab:main}
\end{table}

First, every pullback metric exceeds both model-agnostic baselines in every column by a margin that is many times the standard error. Euclidean and cosine distance are nearly indistinguishable from each other, which is expected since the variants of a family have similar norms, and both sit at a Spearman correlation of about 0.21 on Qwen3 and 0.15 on GPT-2. The weakest pullback entry, the one-sided Euclidean pullback on GPT-2, already reaches 0.35. At the level of individual families the Fisher direct distance exceeds Euclidean distance for 94.5\% of the Qwen3 families and 99.5\% of the GPT-2 families, and the Fisher straight-path length does so for every family of both models.

Second, the three constructions are strictly ordered for every pullback metric and both models, with one-sided below direct below straight path. Evaluating the metric at the edited prompt as well as at the base prompt raises the Fisher Spearman correlation on Qwen3 from 0.40 to 0.56, and integrating it along the straight path raises it further to 0.84. The corresponding GPT-2 values are 0.40, 0.50 and 0.91. The ordering shows that the metric varies appreciably over the segment between two prompts even for single-token edits, a point studied further in Section~\ref{sec:geodesic}, and that the Riemannian picture in which output movement is the integrated local metric reproduces the realised output change closely.

Third, the Fisher pullback is the best metric in every column for Qwen3 and in every column but one for GPT-2, where the angular pullback edges it on the direct Spearman correlation, 0.531 against 0.504, and ties it on the direct Pearson correlation. In Spearman correlation the advantage of the Fisher metric over the Euclidean pullback is of the order of 0.03 to 0.05 in every mode and on both models, whereas in Pearson correlation it widens in the straight path mode, reaching 0.03 on Qwen3 and 0.06 on GPT-2. This ranking is what Eq.~\eqref{eq:kl-expansion} predicts, since the Fisher pullback is the metric whose length is calibrated to the divergence used as ground truth.

Because every distance is evaluated on the same 200 families, these differences are to be judged by a paired test rather than by the overlap of the standard errors, which is the more conservative comparison. A Wilcoxon signed-rank test over the families confirms the advantage of the Fisher pullback over the Euclidean pullback in Spearman correlation in all three modes and on both models: on Qwen3 $p = 1.5\times10^{-19}$, $4.7\times10^{-16}$ and $1.6\times10^{-25}$ for the one-sided, direct and straight path modes, and on GPT-2 $p = 3.3\times10^{-24}$, $3.0\times10^{-8}$ and $1.8\times10^{-33}$, the Fisher metric being the better of the two on between 66\% and 96\% of individual families, the weakest case being the direct mode on GPT-2 and the strongest the straight path mode on the same model. In Pearson correlation the margin is equally clear in the straight path mode, $p < 10^{-13}$ on both models, but it is marginal in the one-sided and direct modes on Qwen3, $p = 0.048$ and $p = 0.041$ against $\sqrt{\sKL}$ and not significant against raw $\sKL$, $p = 0.54$ and $p = 0.52$, so that in those two cells of Table~\ref{tab:rawpmcc} the Fisher and Euclidean pullbacks are to be regarded as tied. The one departure from the ranking is itself significant, on GPT-2 the angular pullback exceeds the Fisher pullback in the direct mode, $p = 6\times10^{-5}$, while tying it in Pearson correlation, $p = 0.79$. The secondary analysis with the hidden state as output, reported in Appendix~\ref{app:tables}, shows the same qualitative picture with the Euclidean pullback on $h$ as the leading metric against $\|h_j - h_i\|$, which again matches the output metric to the ground truth.

The interpretation of these results is that single-token rewordings, which are nearly equidistant in input space, are amplified very unequally by the model. The Jacobian captures this direction-dependent amplification, and the pullback metric converts it into a distance that tracks the realised change in the output distribution. The dependence of the result on the construction of the pairwise distance indicates that most of the information about a specific edit resides at and between the two prompts rather than at the base prompt alone, which has consequences for prediction that are examined in Section~\ref{sec:prediction}.

\subsection{Refining the straight path approximation}
\label{sec:geodesic}

The straight path column of Table~\ref{tab:main} uses five segments. Whether this discretisation is adequate depends on how the metric varies along the straight path, and Table~\ref{tab:profile} shows the median ratio of the Fisher norm at each node to its mean value at the two endpoints. On GPT-2 the metric is nearly four times larger around the midpoint of the segment than at the ends, while on Qwen3 the increase is milder, with a peak of about 1.6 just beyond the midpoint. An interpolated embedding $x_i + t\delta$ with $0 < t < 1$ is not the embedding of any token, and the models are more sensitive at such off-manifold points than at genuine prompts, GPT-2 markedly so. The five-segment rule places interior nodes at $t = 0.2, 0.4, 0.6, 0.8$, and so samples the inflated region, two of those nodes reaching ratios of 3.80 and 4.15 on GPT-2. What escapes it is the apex and the curvature of the profile around it, and the $K = 10$ row indicates how much of the shape falls between the coarser nodes.

\begin{table}[t]
\centering
\small
\begin{tabular}{l ccccccccccc}
\toprule
$t$ & 0 & 0.1 & 0.2 & 0.3 & 0.4 & 0.5 & 0.6 & 0.7 & 0.8 & 0.9 & 1 \\
\midrule
Qwen3-0.6B, $K=10$ & 0.95 & 1.06 & 1.22 & 1.34 & 1.49 & 1.58 & 1.64 & 1.55 & 1.40 & 1.19 & 1.05 \\
Qwen3-0.6B, $K=5$  & 0.96 & & 1.22 & & 1.43 & & 1.57 & & 1.36 & & 1.04 \\
GPT-2-small, $K=5$ & 0.95 & & 1.40 & & 3.80 & & 4.15 & & 1.58 & & 1.05 \\
\bottomrule
\end{tabular}
\caption{Median over all pairs of the Fisher pullback norm $\|\delta\|_{F}$ at the node $x_i + t\delta$, divided by its mean value at the two endpoints. The $K = 10$ row uses the 40-family Qwen3 subset and the $K = 5$ rows use all 200 families of each model.}
\label{tab:profile}
\end{table}

To quantify the effect, the path integral was recomputed with $K = 10$ segments on a subset of 40 Qwen3 families, taken as every fifth family, and the trapezoid rule was then evaluated on the same nodes for $K = 1$, $2$, $5$ and $10$. Table~\ref{tab:kconv} reports the result for the three pullback metrics on the distribution. The correlations increase monotonically with $K$ for every metric. For the Fisher pullback the Spearman correlation rises from 0.56 at $K = 1$, which is the direct distance, through 0.83 at $K = 5$ to 0.87 at $K = 10$, and the Pearson correlation rises from 0.41 through 0.78 to 0.87, so that at $K = 10$ the integrated Fisher length is linearly, and not merely monotonically, related to $\sqrt{\sKL}$. The median relative change of an individual Fisher length between $K = 5$ and $K = 10$ is 6\%. The ranking of the three metrics is preserved at every $K$, with Fisher ahead of angular ahead of Euclidean. The gains between $K = 5$ and $K = 10$ are smaller than those between $K = 2$ and $K = 5$ but have not vanished, so the $K = 10$ values should be regarded as close to, rather than at, convergence. A check on three GPT-2 families, where the mid-path peak is sharper, gave Fisher Spearman and Pearson values of 0.50 and 0.42 at $K = 1$, 0.93 and 0.78 at $K = 5$ and 0.95 and 0.97 at $K = 10$, so the same conclusion applies with greater force to that model.

\begin{table}[t]
\centering
\footnotesize
\setlength{\tabcolsep}{3.5pt}
\begin{tabular}{l cc cc cc}
\toprule
 & \multicolumn{2}{c}{Pullback Euclidean} & \multicolumn{2}{c}{Pullback angular} & \multicolumn{2}{c}{Pullback Fisher} \\
\cmidrule(lr){2-3}\cmidrule(lr){4-5}\cmidrule(lr){6-7}
$K$ & $\rho$ & $r$ & $\rho$ & $r$ & $\rho$ & $r$ \\
\midrule
1 (direct) & $0.515 \pm 0.025$ & $0.399 \pm 0.024$ & $0.521 \pm 0.024$ & $0.393 \pm 0.026$ & $0.557 \pm 0.022$ & $0.411 \pm 0.022$ \\
2          & $0.591 \pm 0.025$ & $0.480 \pm 0.025$ & $0.609 \pm 0.025$ & $0.487 \pm 0.024$ & $0.631 \pm 0.023$ & $0.521 \pm 0.025$ \\
5          & $0.794 \pm 0.017$ & $0.743 \pm 0.022$ & $0.808 \pm 0.015$ & $0.758 \pm 0.019$ & $0.828 \pm 0.015$ & $0.782 \pm 0.019$ \\
10         & $0.838 \pm 0.015$ & $0.831 \pm 0.015$ & $0.852 \pm 0.012$ & $0.842 \pm 0.011$ & $\mathbf{0.867 \pm 0.013}$ & $\mathbf{0.865 \pm 0.011}$ \\
\bottomrule
\end{tabular}
\caption{Convergence of the straight-path length with the number of path segments $K$ on 40 Qwen3-0.6B families. Entries are the mean within-family Spearman correlation with $\sKL$ and Pearson correlation with $\sqrt{\sKL}$, with standard errors over families. All rows are evaluated on the same eleven nodes per pair.}
\label{tab:kconv}
\end{table}

Two remarks follow. The quantity computed is the length of the straight segment in input space and not the geodesic distance proper, which is the infimum of lengths over all paths and which a straight segment through a region of inflated metric would overestimate. The improvement with $K$ nonetheless shows that the local metric, integrated faithfully along even this simple path, reproduces the realised divergence with a Pearson correlation close to 0.9. Related local and path-based Fisher quantities have been compared in parameter-space pruning \cite{Berman:2026rvt}; here increasing $K$ refines integration along a fixed path rather than optimising the path. The second remark concerns cost. The direct and straight-path lengths require forward-mode evaluations at both prompts and at the interior nodes, whereas the ground truth requires a single ordinary forward pass per prompt. Their value therefore lies in demonstrating that the Riemannian picture is correct and in the geometric information that the metric carries, rather than in computational savings, which can only come from the one-sided construction.

%%%%%%%%%%%%%%%%%%%%%%%%%%%%%%%%%%%%%%%%%%%%%%%%%
\section{Applications to Interpretability}
\label{sec:applications}

Section~\ref{sec:results} established that the pullback metrics measure the effect of a rewording far better than the model-agnostic distances, and that the Fisher pullback performs best in all but one of the comparisons of Table~\ref{tab:main}. This section therefore restricts attention to the Fisher distances and asks what the geometry they induce reveals about the behaviour of the model. Three questions are addressed in turn, namely whether the effect of an edit can be anticipated from the base prompt alone, how the quality of the metric depends on the type of message being reworded, and which kinds of edit the geometry identifies as the most disruptive. These are the questions that arise when a deployed assistant is tested for robustness to the wording of its inputs, and Section~\ref{sec:uses} draws the resulting uses together.

\subsection{Predicting the effect of an edit from the base prompt}
\label{sec:prediction}

The question for a practical chatbot use is whether the metric evaluated once at a customer message can rank the possible rewordings of that message by the disruption they would cause, without the edited messages ever being run. This is the one-sided construction restricted to the 50 pairs that join each seed to its variants. Table~\ref{tab:seededit} reports the mean within-family Spearman correlation between the predicted and the realised reply shift $\sqrt{\sKL}$ on these pairs, and it separates the two endpoints so that the metric evaluated at the seed can be compared with the metric evaluated at the edit.

\begin{table}[t]
\centering
\small
\begin{tabular}{l ccccc}
\toprule
 & Euclidean & Fisher at seed & Fisher at edit & Fisher direct & Fisher straight path \\
\midrule
Qwen3-0.6B  & 0.214 & 0.218 & 0.608 & 0.635 & 0.899 \\
GPT-2-small & 0.064 & 0.354 & 0.434 & 0.474 & 0.940 \\
\bottomrule
\end{tabular}
\caption{Mean within-family Spearman correlation between an input distance and the realised reply shift $\sqrt{\sKL}$ over the 50 seed-to-variant pairs of each family. The Fisher metric is evaluated at the seed only, at the edited prompt only, averaged over the two, or integrated along the path with $K = 5$.}
\label{tab:seededit}
\end{table}

On GPT-2 the metric at the seed alone is informative. It attains a Spearman correlation of 0.35 against 0.06 for Euclidean distance, it exceeds Euclidean distance on 86\% of the families, and of the five variants it flags as most disruptive an average of 1.39 are truly among the five most disruptive, against 0.5 expected by chance. On Qwen3 the same quantity attains 0.22, which is statistically indistinguishable from the 0.21 of Euclidean distance, exceeds it on only 52\% of the families and recovers 0.77 of the top five. The metric evaluated at the edited prompt, by contrast, reaches 0.61 on Qwen3, so the information that the one-sided construction lacks is present in the geometry at the other end of the segment. Two features of the data bear on this asymmetry. The mid-path growth of the metric in Table~\ref{tab:profile} shows that a full token substitution carries the input beyond the radius within which the linearisation at the base prompt is accurate, and the base-prompt ruler systematically misranks one class of edits, as shown in Section~\ref{sec:categories}. Euclidean distance is also already more informative for Qwen3 than for GPT-2, at 0.21 against 0.06, so that the input embeddings of the instruction-tuned model encode more of the relevant structure and leave less for the first-order correction to add. 

In summary, the cheapest prediction mode is useful for one of the two models, while the more expensive modes track the realised effect of these edits substantially better on both.

\subsection{Dependence on message type}
\label{sec:types}

Table~\ref{tab:groups} breaks the Fisher results down by the group of the seed message. For the pairwise constructions the quality of the metric is uniform across groups, with every group lying between 0.82 and 0.86 in straight path Spearman correlation on Qwen3 and between 0.90 and 0.93 on GPT-2, and with the stable messages of Qwen3 being the best described by the direct distance at 0.64. The quality rises slightly with message length on Qwen3, with a Spearman correlation between family length and family quality of about 0.24 for all three constructions, and it falls for families whose outputs move the most, with correlations between the median $\sqrt{\sKL}$ of a family and its quality of $-0.22$, $-0.33$ and $-0.16$ for the one-sided, direct and straight path constructions. Both trends are consistent with a first-order description that is most accurate where the output moves least.

\begin{table}[t]
\centering
\footnotesize
\setlength{\tabcolsep}{3.5pt}
\begin{tabular}{l r ccc ccc c}
\toprule
 & & \multicolumn{3}{c}{Fisher Spearman $\rho$, all pairs} & \multicolumn{3}{c}{Seed-to-edit $\rho$} & \\
\cmidrule(lr){3-5}\cmidrule(lr){6-8}
Group & $n$ & one-sided & direct & straight path & Euclidean & Fisher one-sided & Fisher straight path & Median $\sqrt{\sKL}$ \\
\midrule
\multicolumn{9}{l}{\emph{Qwen3-0.6B}} \\
adversarial   & 17 & 0.381 & 0.541 & 0.833 & 0.121 & 0.317 & 0.884 & 0.447 \\
known fragile & 29 & 0.388 & 0.572 & 0.840 & 0.135 & 0.107 & 0.898 & 0.427 \\
minimal edit  & 57 & 0.371 & 0.534 & 0.824 & 0.238 & 0.175 & 0.905 & 0.519 \\
paraphrase    & 68 & 0.401 & 0.560 & 0.835 & 0.235 & 0.255 & 0.897 & 0.474 \\
stable        & 29 & 0.468 & 0.635 & 0.862 & 0.249 & 0.263 & 0.904 & 0.480 \\
\midrule
\multicolumn{9}{l}{\emph{GPT-2-small}} \\
adversarial   & 17 & 0.418 & 0.528 & 0.932 & $-0.071$ & 0.321 & 0.952 & 0.266 \\
known fragile & 29 & 0.412 & 0.487 & 0.908 & 0.045 & 0.381 & 0.929 & 0.359 \\
minimal edit  & 57 & 0.405 & 0.483 & 0.898 & 0.087 & 0.381 & 0.936 & 0.500 \\
paraphrase    & 68 & 0.406 & 0.522 & 0.905 & 0.070 & 0.382 & 0.941 & 0.429 \\
stable        & 29 & 0.355 & 0.503 & 0.912 & 0.105 & 0.228 & 0.945 & 0.372 \\
\bottomrule
\end{tabular}
\caption{Fisher pullback results by group of the seed message. The first three numeric columns are the mean within-family Spearman correlations of Table~\ref{tab:main} restricted to the families of each group. The seed-to-edit columns are the correlations of Table~\ref{tab:seededit} for Euclidean distance, which serves as the baseline, and for the Fisher metric at the seed and for the Fisher straight-path length, and the last column is the median realised reply shift within the group. Only the first of those two Fisher columns is available without running the edited messages, so that both are given: the straight path column is the better description of the data and the one-sided column a quicker candidate for a prediction.}
\label{tab:groups}
\end{table}

The same uniformity holds on the seed-to-edit pairs of Section~\ref{sec:prediction}. There the Fisher straight-path length attains a Spearman correlation between 0.88 and 0.91 in every group on Qwen3 and between 0.93 and 0.95 on GPT-2, against at most 0.25 and 0.11 for Euclidean distance, and it exceeds Euclidean distance on every one of the 200 families of both models. The margin is widest for the adversarial and known-fragile messages, at 0.76 for both groups on Qwen3 and at 1.02 and 0.88 on GPT-2, because these are the two groups for which Euclidean distance is least informative on both models. The adversarial group comprises social-engineering attempts, prompt injections and out-of-scope requests, and the known-fragile group comprises messages on which chatbots are empirically unreliable. These are therefore the messages for which the operator of an assistant, here a bank, most needs to know which rewordings would change the behaviour of its assistant, and for which a model-agnostic distance is the poorest guide. The metric evaluated at the seed alone is more group dependent. Its gain over Euclidean distance is again largest on the adversarial messages, at 0.32 against 0.12 on Qwen3 and 0.32 against $-0.07$ on GPT-2, although the Qwen3 result rests on 17 families. On the four remaining Qwen3 groups it adds nothing, and on two of them it is the weaker measure: 0.11 against 0.14 on the known-fragile messages and 0.18 against 0.24 on the minimal-edit ones, with the paraphrase and stable groups level at 0.26 against 0.24 and 0.25 respectively. Adversarial messages are therefore the only Qwen3 group on which the base-prompt ruler carries information that Euclidean distance does not, whereas on GPT-2 it leads on all five groups. Any use of the one-sided construction for triage would have to be restricted accordingly. The limitation is therefore specific to evaluating the geometry at the base prompt: when the Fisher metric is followed along the path to the edited prompt, the resulting straight-path length substantially outperforms Euclidean distance in every message group for both models. Figure~\ref{fig:seeds} displays the Qwen3 breakdown, and the GPT-2 counterpart is given in Appendix~\ref{app:perseed}.

\begin{figure}[t]
\centering
\begin{subfigure}[t]{0.575\textwidth}
\centering
\includegraphics[width=\textwidth]{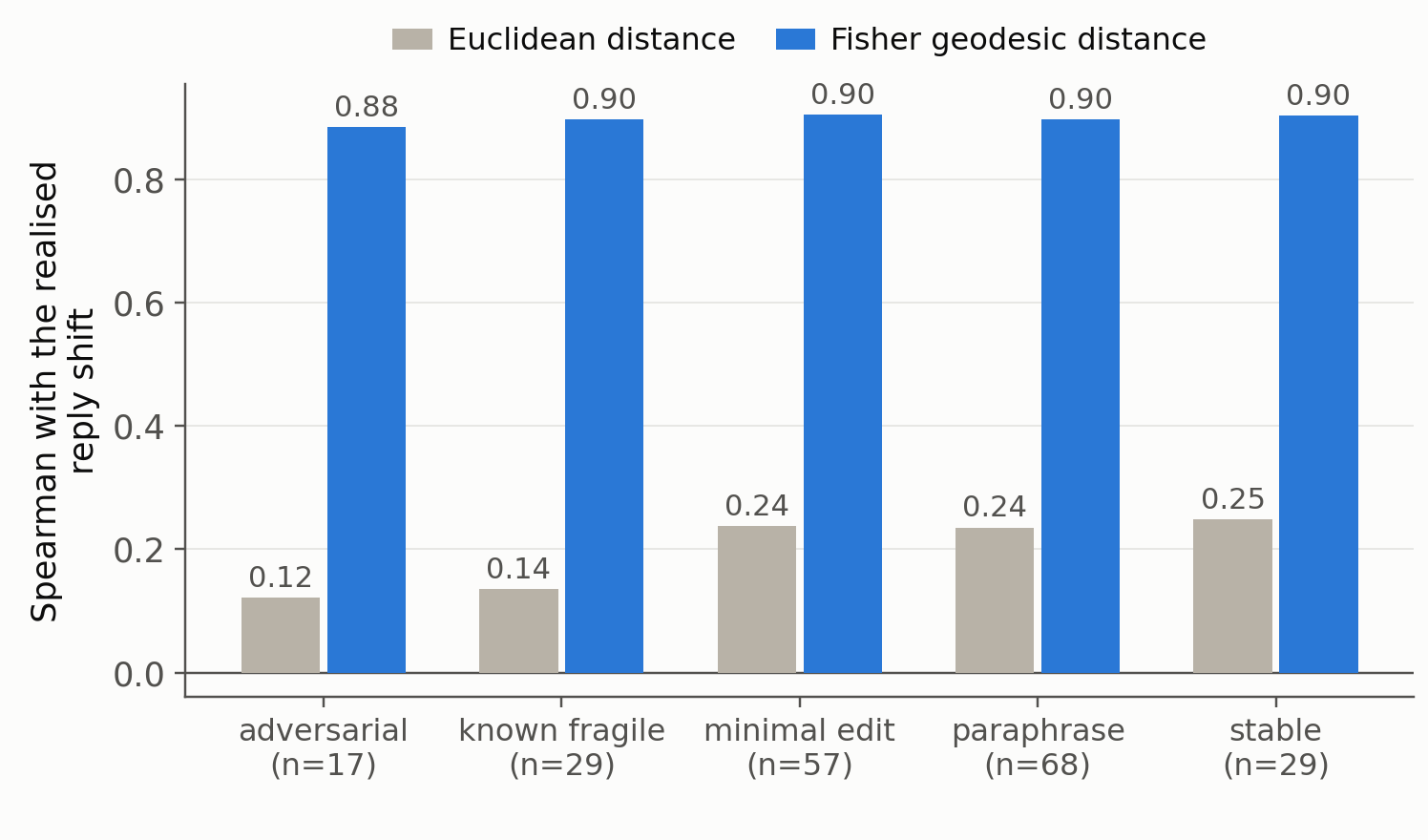}
\caption{Seed-to-edit correlation by message type.}
\label{fig:seeds-groups}
\end{subfigure}\hfill
\begin{subfigure}[t]{0.405\textwidth}
\centering
\includegraphics[width=\textwidth]{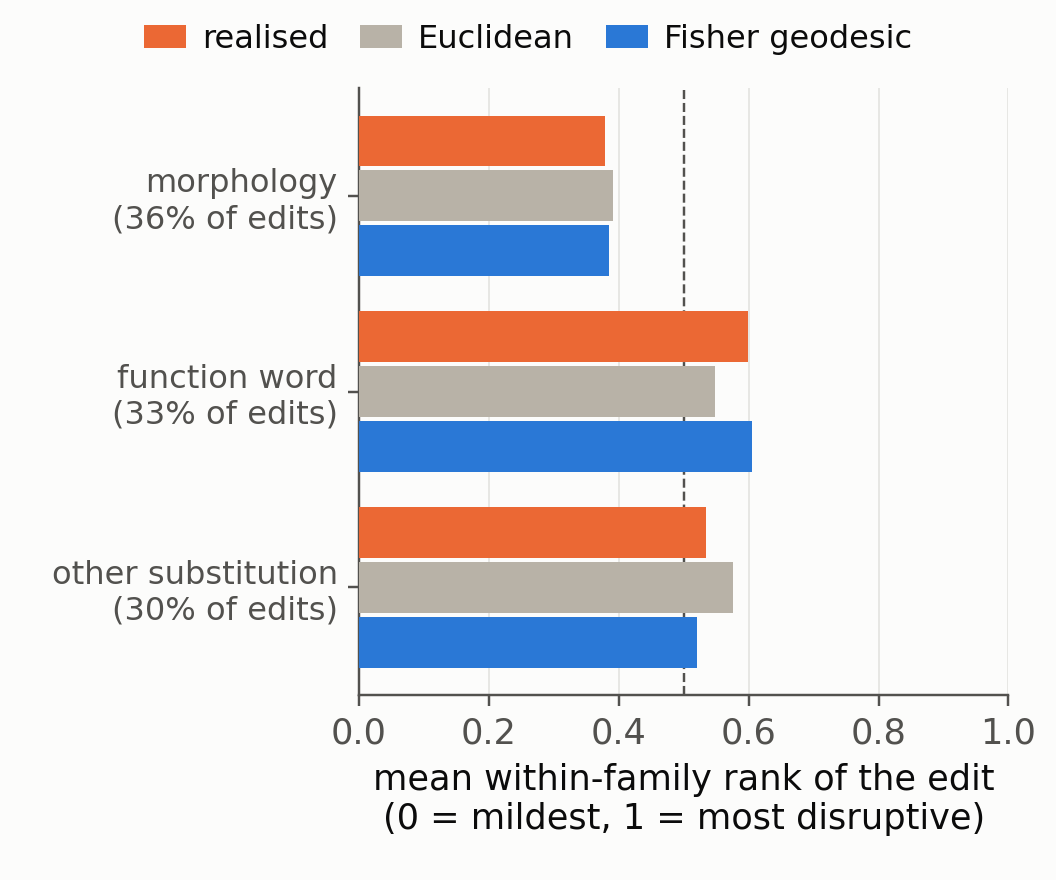}
\caption{Disruptiveness by edit category.}
\label{fig:seeds-categories}
\end{subfigure}
\caption{Per-family analysis for Qwen3-0.6B. Panel (a) gives the mean within-family Spearman correlation with the realised reply shift over the seed-to-variant pairs, by message type, for Euclidean distance and for the Fisher straight-path length. Panel (b) gives the mean within-family rank of the realised reply shift of each edit category, with 0 the mildest and 1 the most disruptive edit of a family, together with the mean ranks assigned to the edits by Euclidean distance and by the Fisher straight-path length.}
\label{fig:seeds}
\end{figure}

\subsection{Dependence on edit category}
\label{sec:categories}

Table~\ref{tab:categories} classifies the 10,000 seed-to-variant substitutions of each model into the three edit categories of Section~\ref{sec:data} and reports, for each category, the median realised reply shift together with the mean within-family rank of the realised shift and the mean ranks assigned to the edits by Euclidean distance and by the Fisher straight-path length. Function-word substitutions, such as a change of preposition, pronoun or determiner, are the most disruptive class on both models, with a median $\sqrt{\sKL}$ of 0.40 on Qwen3 against 0.21 for morphological changes and 0.34 for other substitutions, and 0.27 against 0.20 and 0.22 on GPT-2. Morphological changes such as a change of number or inflection are the least disruptive. This agrees with linguistic intuition, since a function word can change the direction of a transfer, the owner of an account or the polarity of a request, while an inflectional change rarely alters what is being asked.

\begin{table}[t]
\centering
\footnotesize
\setlength{\tabcolsep}{3.5pt}
\begin{tabular}{l rcccc rcccc}
\toprule
 & \multicolumn{5}{c}{Qwen3-0.6B} & \multicolumn{5}{c}{GPT-2-small} \\
\cmidrule(lr){2-6}\cmidrule(lr){7-11}
 & & & \multicolumn{3}{c}{Mean rank} & & & \multicolumn{3}{c}{Mean rank} \\
\cmidrule(lr){4-6}\cmidrule(lr){9-11}
Edit category & Share & \begin{tabular}[b]{@{}c@{}}Median\\$\sqrt{\sKL}$\end{tabular} & realised & Euclidean & \begin{tabular}[b]{@{}c@{}}Fisher\\straight path\end{tabular} & Share & \begin{tabular}[b]{@{}c@{}}Median\\$\sqrt{\sKL}$\end{tabular} & realised & Euclidean & \begin{tabular}[b]{@{}c@{}}Fisher\\straight path\end{tabular} \\
\midrule
morphology         & 36\% & 0.210 & 0.38 & 0.39 & 0.39 & 24\% & 0.199 & 0.46 & 0.58 & 0.45 \\
function word      & 33\% & 0.404 & 0.60 & 0.55 & 0.61 & 38\% & 0.270 & 0.55 & 0.30 & 0.56 \\
other substitution & 30\% & 0.342 & 0.53 & 0.58 & 0.52 & 39\% & 0.221 & 0.48 & 0.64 & 0.47 \\
\bottomrule
\end{tabular}
\caption{Edit categories of the seed-to-variant substitutions. The realised rank is the mean within-family rank of the realised reply shift, normalised so that 0 is the mildest and 1 the most disruptive edit of a family, and the Euclidean and Fisher straight path columns give the same quantity for the ranking of the edits by each distance.}
\label{tab:categories}
\end{table}

The same table shows that the Fisher straight-path length recovers this ordering from the geometry alone. On both models the mean rank that it assigns to each category lies within 0.015 of the realised rank, so that function-word substitutions are identified as the most disruptive class and morphological changes as the least without any linguistic categorisation of the edits. Euclidean distance does not recover the ordering. On GPT-2 it places function-word substitutions at a mean rank of 0.30, below both other categories, where their realised rank is 0.55, and on Qwen3 it under-ranks them and over-ranks the other substitutions. The metric evaluated at the seed alone makes the first of these errors on both models, placing function-word substitutions at a mean rank of about 0.4 against a realised rank of 0.55 to 0.60, since the linearisation at the base prompt treats a function word as a direction of modest sensitivity whereas the realised effect of replacing it is large. This misranking accounts for a substantial part of the shortfall of the one-sided prediction on Qwen3, and its absence from the straight path ranking shows that the missing information is recovered once the metric is followed along the path to the edit.

A single family illustrates the pattern. For the Qwen3 seed \emph{``someone stole my card can you shut it down''} the model replies in English with an apology and a request for details. Replacing \emph{you} by \emph{Vous}, a nearest neighbour of \emph{you} in the multilingual embedding matrix, produces a reply entirely in French, and this edit has the largest realised shift of all 50 variants with $\sqrt{\sKL} = 1.99$. The lowercase variant \emph{vous} behaves in the same way and has the second largest shift, at 1.95. The effect is not particular to French: the next two shifts of the family are \emph{can} to \emph{pueden} at 1.28 and \emph{my} to \emph{meine} at 1.21, which switch the reply into Spanish and German. The four most disruptive edits of this family are therefore all substitutions of an English function word by a translation of it, which the multilingual embedding matrix places among its nearest neighbours. The system prompt instructs the model to answer in the language of the customer, so that one such token is enough to change which language that is taken to be, and these edits move the reply further than any change of content word in the family. The Fisher straight-path length ranks these two edits second and first, whereas Euclidean distance ranks them 32nd and 44th and the metric at the seed 26th and 42nd. At the other extreme the inflectional edits \emph{shut} to \emph{shuts} and \emph{down} to \emph{downs} leave the reply essentially unchanged, with shifts of 0.045 and 0.038, and the straight-path length places the first of these last of all 50 variants and the second in the lower half. Over the 50 variants of the family the straight-path length attains a Spearman correlation of 0.88 with the realised shift, against $-0.17$ for Euclidean distance and $-0.39$ for the metric at the seed. The example shows what the geometry captures when the metric is integrated along the path to the edit, and what the linearisation at the base prompt misses.

\subsection{Uses and extensions}
\label{sec:uses}

The results support three uses of the induced geometry in the quality assurance of a deployed assistant, the cost of which differs with the construction and is reported here alongside its accuracy. Scoring one candidate rewording takes one forward-mode pass in the one-sided construction, two in the direct one and $K+1$, here six, in the straight path one, against the single ordinary forward pass that yields the exact answer. The latter two evaluate the metric at the edited message and so require it to be run; the one-sided construction alone leaves the candidate unevaluated by the model, and it is cheaper than the exact answer only once its cost is amortised, by forming $g_x$ at the original message as described at the end of this section and scoring candidates by a quadratic form.

The first is the ranking of rewordings. Given a customer message and a set of candidate rewordings, the Fisher straight-path length ranks the candidates by the disruption they cause with a Spearman correlation of 0.90 on Qwen3 and 0.94 on GPT-2 against the realised divergence, against 0.21 and 0.06 for Euclidean distance, and it does so more faithfully than Euclidean distance for every message family of both models. Its advantage is largest for messages of the adversarial and known-fragile types, on which a model-agnostic distance is least informative and where red-teaming effort is concentrated. Read as a procedure the comparison is not an even one, since the straight-path length evaluates the candidate rewordings and Euclidean distance does not, and a practitioner able to evaluate them could measure $\sKL$ directly and obtain the ranking exactly. The comparison establishes instead that the integrated local metric accounts for the realised change almost completely, so that the ranking is a property of the geometry rather than of the particular wordings, and that the geometry is therefore the object in which a cheaper predictor is to be sought. The candidate for that role is the one-sided construction, which never runs the candidate rewording, and which reaches 0.35 on GPT-2 against 0.06 for Euclidean distance but only 0.22 against 0.21 on Qwen3.

The second use is the prioritisation of test suites. Function-word variants, including negations, prepositions and possessives, are the most disruptive class of single-token edit and should be over-represented in the regression tests of an intent classifier or a chatbot, while inflectional variants may be sampled sparsely. The straight-path length identifies this class without any linguistic annotation of the edits, so that the same prioritisation can be derived for edit types that have not been categorised in advance.
The third is the selection of robust templates. Among candidate phrasings of a routine request, those at which the pullback metric is small in the directions spanned by plausible customer edits will produce replies that are stable under rephrasing, and the metric offers a quantitative criterion for that choice.

Beyond these uses the geometry is itself an object of interpretability. The kernel of $g_x$ identifies the input perturbations to which the model is locally indifferent. Its leading eigenvectors, when resolved by token position and embedding direction, indicate the perturbation directions which most change the metric and hence the response of the model. 

Several extensions suggest themselves. The ground truth here is the first token of the reply, and extending the output to the distribution over the first several reply tokens would test how well the metric tracks the reply as a whole. The gap between the one-sided and the direct constructions motivates evaluating the metric at the midpoint of the segment or including a second-order term, and forming the full matrix $g_x$ at the base prompt once would allow every single-token substitution at every position to be scored by a single matrix product. The straight path could be replaced by an optimised path, giving the geodesic distance proper and, as a by-product, the minimal edit that achieves a prescribed change in the output. The restriction to equal-length pairs could be lifted by a suitable alignment of token positions. The low effective rank of pullback metrics on transformers \cite{wang2026fishback} suggests that randomised low-rank sketches of $g_x$ could make the construction tractable for larger models, and the variation of the metric along the path is the first step towards the curvature of the induced geometry, which has begun to be studied on the output side of language models \cite{hosseini2023straighten, king2026curvature} but not on the input side.

%%%%%%%%%%%%%%%%%%%%%%%%%%%%%%%%%%%%%%%%%%%%%%%%%
\section{Conclusion}
\label{sec:conclusion}

A frozen language model induces a geometry on the space of its prompts, and this work has shown that the geometry is informative about the behaviour of the model. By pulling the Euclidean, angular, and Fisher metrics of the next-token distribution back along the map from the embeddings of a user message, three model-intrinsic distances between prompts were constructed and compared with the native Euclidean and cosine distance on a bank-chatbot dataset of 10,200 prompts for each of Qwen3-0.6B and GPT-2-small.
Every pullback metric tracks the realised divergence between next-token distributions far better than the model-agnostic baselines, the Fisher pullback is the best or joint best metric in all but one setting, and the three constructions of a pairwise distance are strictly ordered, with the integrated length along the straight path between two prompts reaching a Spearman correlation of 0.84 on Qwen3 and 0.91 on GPT-2 against 0.21 and 0.15 for Euclidean distance. A finer discretisation of the path raises the Qwen3 Fisher correlations to 0.87 in both Spearman and Pearson measures, so that the integrated local metric is close to linearly related to the square root of the realised divergence. These results hold for an instruction-tuned model of current design and for a base model of an earlier generation, which supports their being properties of the prompt map of decoder-only transformers rather than of a single model.

The analysis by message type and edit category shows what the best of these distances, the Fisher straight-path length, delivers in practice. Measured on the edits of a single message it exceeds Euclidean distance on every one of the 200 message families of both models, with its widest margin on the adversarial and known-fragile messages, for which a model-agnostic distance is least informative and the behaviour of a deployed assistant matters most. It also recovers from the geometry alone the ordering of edit types, ranking function-word substitutions as the most disruptive single-token edits on both models and inflectional changes as the least, an ordering that Euclidean distance on GPT-2 and the metric at the base prompt on both models fail to reproduce, since both rank function-word substitutions as the mildest class. The cheapest construction, the metric at the base prompt, is a useful predictor of edit effects on GPT-2 and not yet on Qwen3, and the asymmetry between the two ends of a segment shows that the information it lacks is geometric rather than absent, since the straight-path length recovers it once the metric is followed along the path to the edit.

Several directions for extension naturally follow. The correlations reported here are for the first token of the reply, and whether they persist for the distribution over later tokens remains to be tested. Given the computational budget for a path search, the straight segment could be replaced by an optimised path, which would give the geodesic distance proper. And the construction should be scaled to larger models through low-rank approximation of the metric. 

The central conclusion is that the geometry pulled back from the output of a language model is a well-defined, computable, and behaviourally validated object on prompt space, and that the Fisher metric is its natural form. It supplies a ruler calibrated to what the model does, and it substantially outperforms the model-agnostic baselines. It therefore provides a quantitative means of studying the sensitivity and stability of prompts.

%%%%%%%%%%%%%%%%%%%%%%%%%%%%%%%%%%%%%%%%%%%%%%%%%
\section*{Acknowledgements}
EH is supported by S\~ao Paulo Research Foundation (FAPESP) grant 2024/18994-7.
TS is supported by FAPESP grant 2022/09891-4.
EH and TS carried out manuscript revisions with the assistance of Claude (Anthropic) and reviewed results and code with the assistance of Codex (OpenAI). EH and TS acknowledge access to OpenAI models through the ChatGPT for Academic Researchers program.

\addcontentsline{toc}{section}{References}
\bibliographystyle{utphys}
\bibliography{references}{}

%%%%%%%%%%%%%%%%%%%%%%%%%%%%%%%%%%%%%%%%%%%%%%%%%
\begin{appendices}

The appendices collect material that supports the results of Sections~\ref{sec:results} and~\ref{sec:applications} without being needed to follow them. Appendix~\ref{app:tables} reports two variants of the main comparison of Table~\ref{tab:main}, one in which the Pearson correlation is computed against the raw symmetric divergence rather than its square root and one in which the final hidden state replaces the next-token distribution as the output of the map. Appendix~\ref{app:perseed} extends the per-family analysis of Section~\ref{sec:applications}, breaking the Fisher results down by the length of the seed message and giving the GPT-2 counterpart of Figure~\ref{fig:seeds}. Appendix~\ref{app:repro} describes the pipeline and the computational set-up needed to reproduce the experiments.

\section{Additional tables}
\label{app:tables}

Table~\ref{tab:rawpmcc} reports the Pearson correlation of each distance against the raw symmetric divergence $\sKL$ rather than its square root. The values are uniformly lower than those against $\sqrt{\sKL}$ in Table~\ref{tab:main}, as expected from the quadratic relation of Eq.~\eqref{eq:kl-expansion}, and the ordering of metrics and constructions is unchanged. Table~\ref{tab:hidden} reports the secondary analysis in which the output is the final hidden state $h$ and the ground truth is $\|h_j - h_i\|$. The Fisher metric has no counterpart here since $h$ is not a distribution. The Euclidean pullback on $h$ is the leading metric, in keeping with the matching of output metric to ground truth, and the ordering of constructions is the same as for the distribution.

\begin{table}[h]
\centering
\footnotesize
\setlength{\tabcolsep}{3pt}
\begin{tabular}{l ccc ccc}
\toprule
 & \multicolumn{3}{c}{Qwen3-0.6B} & \multicolumn{3}{c}{GPT-2-small} \\
\cmidrule(lr){2-4}\cmidrule(lr){5-7}
Metric & one-sided & direct & straight path & one-sided & direct & straight path \\
\midrule
Euclidean          & $0.122 \pm 0.009$ & $0.122 \pm 0.009$ & $0.122 \pm 0.009$ & $0.095 \pm 0.012$ & $0.095 \pm 0.012$ & $0.095 \pm 0.012$ \\
Cosine             & $0.124 \pm 0.009$ & $0.124 \pm 0.009$ & $0.124 \pm 0.009$ & $0.098 \pm 0.013$ & $0.098 \pm 0.013$ & $0.098 \pm 0.013$ \\
Pullback Euclidean & $\mathbf{0.266 \pm 0.010}$ & $\mathbf{0.381 \pm 0.013}$ & $0.717 \pm 0.009$ & $0.244 \pm 0.012$ & $0.353 \pm 0.017$ & $0.740 \pm 0.011$ \\
Pullback angular   & $0.259 \pm 0.011$ & $0.372 \pm 0.015$ & $0.712 \pm 0.008$ & $0.259 \pm 0.012$ & $\mathbf{0.381 \pm 0.016}$ & $0.726 \pm 0.011$ \\
Pullback Fisher    & $\mathbf{0.268 \pm 0.010}$ & $\mathbf{0.385 \pm 0.012}$ & $\mathbf{0.741 \pm 0.008}$ & $\mathbf{0.271 \pm 0.012}$ & $\mathbf{0.382 \pm 0.016}$ & $\mathbf{0.796 \pm 0.008}$ \\
\bottomrule
\end{tabular}
\caption{Mean within-family Pearson correlation between each input-space distance and the raw symmetric Kullback--Leibler divergence, with standard errors over the 200 families. Straight-path lengths use $K = 5$. Bold marks the best entry in each column, with ties shown jointly where a paired test over the families does not separate the two.}
\label{tab:rawpmcc}
\end{table}

\begin{table}[h]
\centering
\footnotesize
\setlength{\tabcolsep}{3pt}
\begin{tabular}{l ccc ccc}
\toprule
 & \multicolumn{3}{c}{Spearman $\rho$} & \multicolumn{3}{c}{Pearson $r$} \\
\cmidrule(lr){2-4}\cmidrule(lr){5-7}
Metric & one-sided & direct & straight path & one-sided & direct & straight path \\
\midrule
\multicolumn{7}{l}{\emph{Qwen3-0.6B}} \\
Euclidean            & $0.215 \pm 0.010$ & $0.215 \pm 0.010$ & $0.215 \pm 0.010$ & $0.184 \pm 0.009$ & $0.184 \pm 0.009$ & $0.184 \pm 0.009$ \\
Cosine               & $0.223 \pm 0.010$ & $0.223 \pm 0.010$ & $0.223 \pm 0.010$ & $0.184 \pm 0.009$ & $0.184 \pm 0.009$ & $0.184 \pm 0.009$ \\
Pullback Euclidean   & $\mathbf{0.462 \pm 0.007}$ & $\mathbf{0.648 \pm 0.009}$ & $\mathbf{0.872 \pm 0.004}$ & $\mathbf{0.369 \pm 0.008}$ & $\mathbf{0.522 \pm 0.010}$ & $\mathbf{0.819 \pm 0.006}$ \\
Pullback angular     & $0.460 \pm 0.007$ & $0.643 \pm 0.009$ & $0.871 \pm 0.004$ & $0.364 \pm 0.008$ & $0.513 \pm 0.010$ & $0.815 \pm 0.006$ \\
\midrule
\multicolumn{7}{l}{\emph{GPT-2-small}} \\
Euclidean            & $0.155 \pm 0.012$ & $0.155 \pm 0.012$ & $0.155 \pm 0.012$ & $0.145 \pm 0.012$ & $0.145 \pm 0.012$ & $0.145 \pm 0.012$ \\
Cosine               & $0.158 \pm 0.012$ & $0.158 \pm 0.012$ & $0.158 \pm 0.012$ & $0.145 \pm 0.012$ & $0.145 \pm 0.012$ & $0.145 \pm 0.012$ \\
Pullback Euclidean   & $0.324 \pm 0.011$ & $0.386 \pm 0.012$ & $\mathbf{0.791 \pm 0.006}$ & $0.312 \pm 0.013$ & $0.404 \pm 0.016$ & $\mathbf{0.775 \pm 0.008}$ \\
Pullback angular     & $\mathbf{0.338 \pm 0.011}$ & $\mathbf{0.389 \pm 0.012}$ & $0.729 \pm 0.007$ & $\mathbf{0.328 \pm 0.013}$ & $\mathbf{0.422 \pm 0.017}$ & $0.709 \pm 0.010$ \\
\bottomrule
\end{tabular}
\caption{Secondary analysis with the final hidden state as output. The pullback metrics are $\|dh\|$ and its angular counterpart, the ground truth is $\|h_j - h_i\|$ for both correlation measures, and straight-path lengths use $K = 5$. Entries are means over the 200 families with standard errors.}
\label{tab:hidden}
\end{table}

\section{Additional per-family results}
\label{app:perseed}

\begin{figure}[tbp]
\centering
\begin{subfigure}[t]{0.45\textwidth}
\centering
\includegraphics[width=\textwidth]{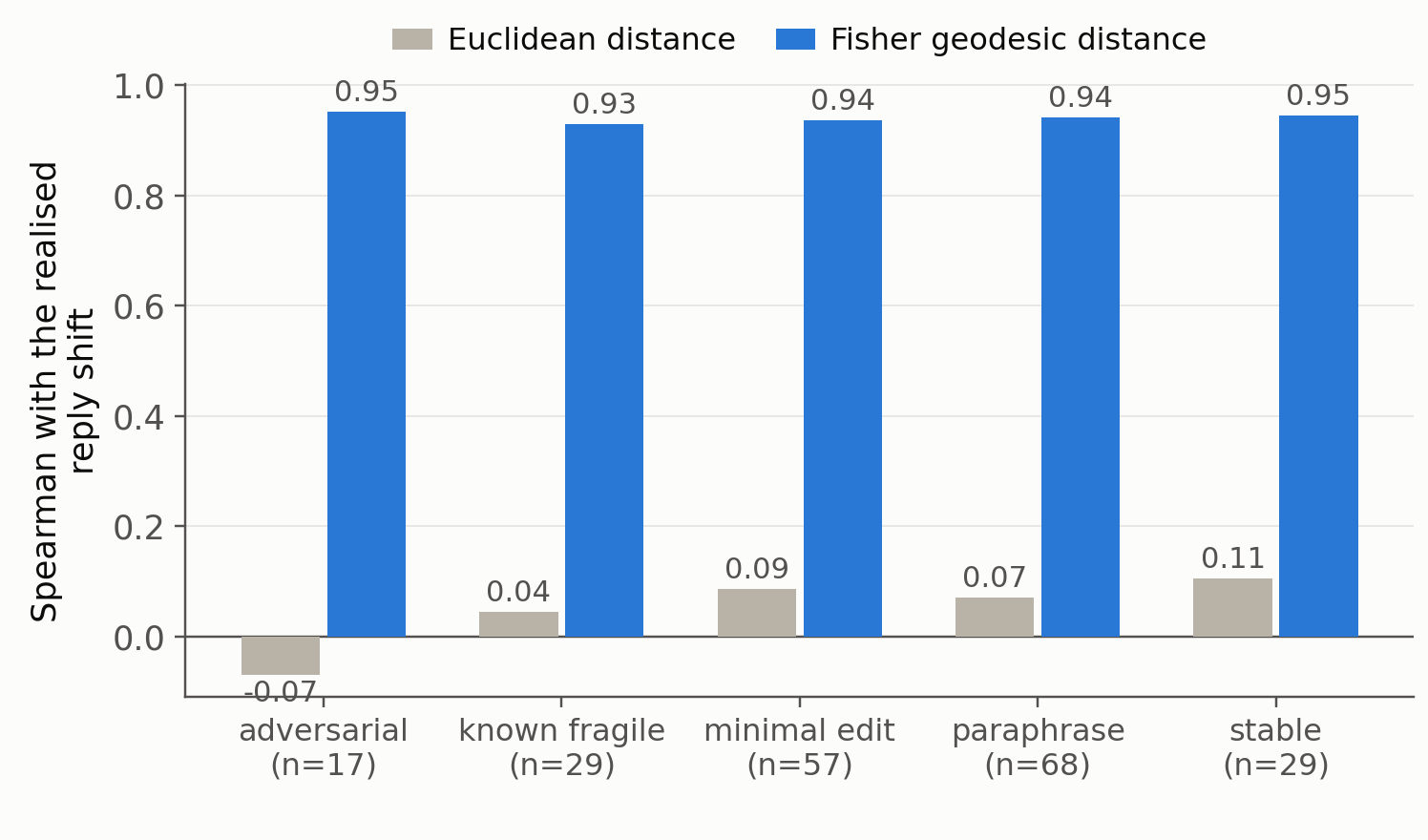}
\caption{Seed-to-edit correlation by message type.}
\label{fig:seeds-gpt2-groups}
\end{subfigure}\hfill
\begin{subfigure}[t]{0.35\textwidth}
\centering
\includegraphics[width=\textwidth]{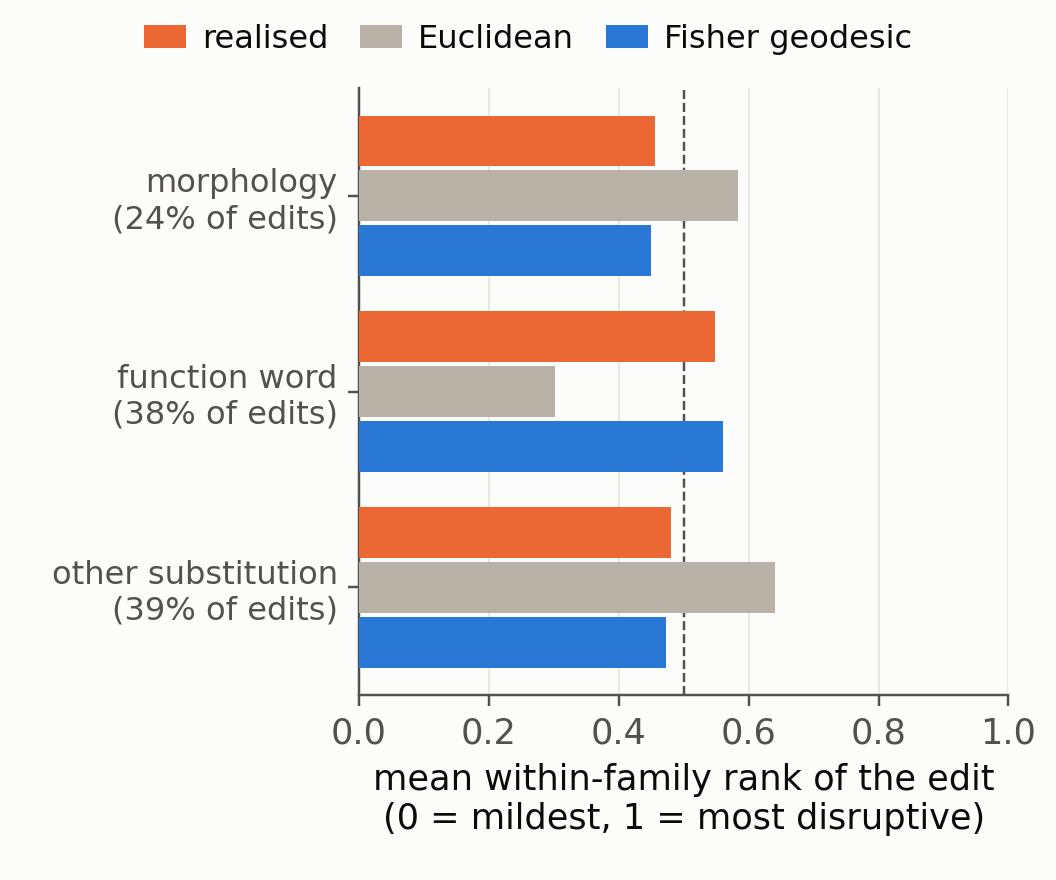}
\caption{Disruptiveness by edit category.}
\label{fig:seeds-gpt2-categories}
\end{subfigure}
\caption{Per-family analysis for GPT-2-small, in the format of Figure~\ref{fig:seeds}.}
\label{fig:seeds-gpt2}
\end{figure}

Table~\ref{tab:length} breaks the Qwen3 and GPT-2 Fisher results down by the token length of the seed message, in tertiles. On both models the short seeds have between 4 and 8 tokens, and the medium seeds 9 or 10, while the long seeds run from 11 up to 17 tokens on Qwen3 (up to 15 on GPT-2), and the three groups are of unequal size because many seeds share the lengths at the tertile boundaries. On Qwen3 all three constructions improve mildly with length while the seed-to-edit prediction does not, and on GPT-2 the one-sided and direct constructions show no length trend, the correlation between family length and family quality being $-0.01$ and $+0.01$, while the straight path construction shows a weak one at $+0.14$ and the seed-to-edit prediction none at $-0.08$. Figure~\ref{fig:seeds-gpt2} shows the per-family analysis of Figure~\ref{fig:seeds} for GPT-2-small, with similar qualitative behaviour.

\begin{table}[t]
\centering
\footnotesize
\setlength{\tabcolsep}{3pt}
\begin{tabular}{l r cccc r cccc}
\toprule
 & \multicolumn{5}{c}{Qwen3-0.6B} & \multicolumn{5}{c}{GPT-2-small} \\
\cmidrule(lr){2-6}\cmidrule(lr){7-11}
Length & $n$ & one-sided & direct & straight path & seed-to-edit & $n$ & one-sided & direct & straight path & seed-to-edit \\
\midrule
short  & 88 & 0.378 & 0.527 & 0.820 & 0.216 & 87 & 0.405 & 0.503 & 0.898 & 0.373 \\
medium & 51 & 0.407 & 0.593 & 0.848 & 0.208 & 57 & 0.385 & 0.499 & 0.916 & 0.314 \\
long   & 61 & 0.421 & 0.592 & 0.850 & 0.227 & 56 & 0.408 & 0.510 & 0.910 & 0.363 \\
\bottomrule
\end{tabular}
\caption{Fisher pullback results by tertile of seed length in tokens. The first three numeric columns of each model are the mean within-family Spearman correlations with $\sKL$ over all pairs and the last is the seed-to-edit correlation of Table~\ref{tab:seededit} for the metric at the seed.}
\label{tab:length}
\end{table}

\section{Reproduction}
\label{app:repro}

The pipeline consists of four scripts. A model wrapper builds the prompt from token identifiers, exposes the embeddings of the user message as the free variable and evaluates the batched forward-mode Jacobian products of Eq.~\eqref{eq:jacobian}. A dataset script constructs the dense local expansion of Section~\ref{sec:data} for a given tokeniser. A distance script evaluates, for every within-family pair and every node of the straight path, the five pullback norms (the three on the distribution and the two on the hidden state) of Section~\ref{sec:maths} together with the ground-truth divergences, and a table script merges the resulting arrays and computes the within-family correlations and their standard errors. The experiments were run with PyTorch 2.5.1 and Transformers 4.51.3 in full single precision with reduced-precision matrix multiplication disabled. The Qwen3 distance computation was sharded over four GPU instances (NVIDIA A10G and L4), at approximately 95 seconds per family on an A10G, and the $K = 10$ convergence study used every fifth family.

\end{appendices}

%%%%%%%%%%%%%%%%%%%%%%%%%%%%%%%%%%%%%%%%%%%%%%%%%

\end{document}